\documentclass{article}
\usepackage{jfrExamplee}
\usepackage[T1]{fontenc}
\usepackage{graphicx}
\usepackage[sort&compress,sectionbib,numbers]{natbib}

\usepackage{setspace}
\usepackage{url}
\usepackage{colortbl}
\usepackage{placeins}
\usepackage{xcolor}
\usepackage{multicol} 
\usepackage{subcaption}
\usepackage{numprint}
\usepackage{breakcites}
\usepackage{amssymb}
\usepackage{wrapfig}
\usepackage{makecell}
\usepackage{lipsum}

\usepackage{url}
\usepackage{eso-pic}

\makeatletter
\newcommand\newtag[2]{#1\def\@currentlabel{#1}\label{#2}}
\makeatother

\usepackage{units}
\usepackage{amsmath} 
\usepackage{enumitem}
\usepackage{hyperref}
\usepackage[acronym,nowarn,hyperfirst=true]{glossaries}

\hypersetup{
    colorlinks,
    linkcolor={blue!30!black},
    citecolor={blue!30!black},
    urlcolor={blue!30!black}
}

\usepackage{acronym}

\newacronym{TI}{TI}{Thermal-Infrared}
\newacronym{LWIR}{LWIR}{Long-Wave Infrared}
\newacronym{UAV}{UAV}{Unmanned Aerial Vehicle}
\newacronym{CNN}{CNN}{Convolutional Neural Network}
\newacronym{HOG}{HOG}{Histogram of Oriented Gradient}
\newacronym{SVM}{SVM}{Support Vector Machine}
\newacronym{FPN}{FPN}{Feature Pyramid Networks}
\newacronym{SSD}{SSD}{Single Shot Detector}
\newacronym{PDF}{PDF}{Probability Density Function}
\newacronym{SR}{SR}{Systematic Resampling}
\newacronym{PF}{PF}{Particle Filter}
\newacronym{UTM}{UTM}{Universal Transverse Mercator}
\newacronym{IMU}{IMU}{Inertial Measurement Unit}
\newacronym{IOU}{IoU}{Intersection over Union}
\newacronym{SAR}{SaR}{Search and Rescue}
\newacronym{DL}{DL}{Deep Learning}

\title{Deep Learning-based Human Detection for UAVs with Optical and Infrared Cameras: System and Experiments}

\author{
Timo Hinzmann\textsuperscript{1} 
\And
Tobias Stegemann\textsuperscript{1,2} 
\And
Cesar Cadena\textsuperscript{1} 
\And
Roland Siegwart\textsuperscript{1}  \\ 
\AND
\textsuperscript{1} Autonomous Systems Lab\\
ETH Zurich\\
Zurich, Switzerland \\
\texttt{firstname.lastname@mavt.ethz.ch} 
\And
\textsuperscript{2} Mantis Technologies\\
Zurich, Switzerland \\
\texttt{tobias@mantistechnologies.ch}
}

\begin{document}

\maketitle

\begin{abstract}
In this paper, we present our deep learning-based human detection system that uses optical (RGB) and long-wave infrared (LWIR) cameras to detect, track, localize, and re-identify humans from UAVs flying at high altitude.
In each spectrum, a customized RetinaNet network with ResNet backbone provides human detections which are subsequently fused to minimize the overall false detection rate.
%
%
We show that by optimizing the bounding box anchors and augmenting the image resolution the number of missed detections from high altitudes can be decreased by over 20 percent.
Our proposed network is compared to different RetinaNet and YOLO variants, and to a classical optical-infrared human detection framework that uses hand-crafted features.
Furthermore, along with the publication of this paper, we release a collection of annotated optical-infrared datasets recorded with different UAVs during search-and-rescue field tests and the source code of the implemented annotation tool.
\end{abstract}


\section{Introduction}
\label{sec:Introduction}

\begin{figure}[htb]
\includegraphics[width=\linewidth, trim=10 0 20 0, clip]{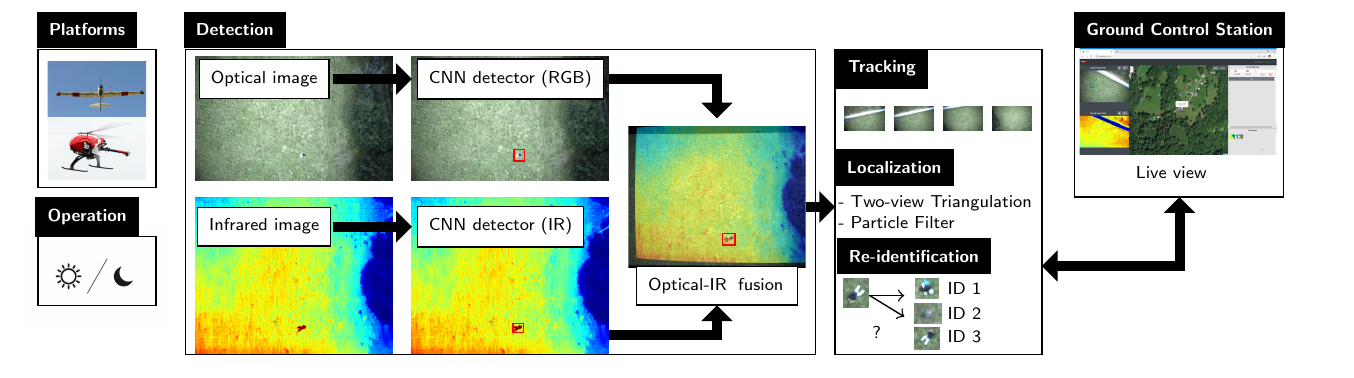}
\caption{Proposed deep learning-based human detection system that uses optical (RGB) and long-wave infrared (LWIR) cameras to detect, track, localize, and re-identify humans from \glspl*{UAV} flying at high altitudes.}
\label{fig:overview}
\end{figure}

The need for robust human detection algorithms is tremendous and has massively increased over the past years due to the vast amount of emerging applications in the field~\cite{Dollar2009,Nguyen2016}.
With \gls*{UAV} technology blooming, research in the field of human detection from aerial views also steadily evolved and experienced much interest for real-world \gls*{SAR} missions~\cite{Rudol2008,Andriluka2010,Blondel2014,Kummerle2016,Bejiga2017}.
While the majority of the earlier work on human detection incorporated hand-crafted features, more recent publications started to make use of \gls*{DL}-based detectors, mostly in the form of \glspl*{CNN}~\cite{DeOliveira2016,Herrmann2016,Bejiga2017}.
In the superordinate field of object detection, deep \glspl*{CNN} have been already established for several years~\cite{Krizhevsky2012,Simonyan2014}, and current state-of-the-art detectors produce impressive results with possible real-time performance~\cite{Redmon2018,Lin2017_RetinaNet}. 
There exist a few publications known to date that try to use these insights from the field of object detection for human detection in aerial images~\cite{Bondi2018,Wang2018,Liu2018,Perera2018,Chang2018,Yun2019}.
%
Making use of either only optical or only \gls*{LWIR} (also referred to as \gls*{TI}) images, still no work so far has considered the use of deep \glspl*{CNN} for combining information from both \gls*{TI} and optical images.
As \glspl*{CNN} need a vast amount of data to outperform hand-crafted detectors, the lack of publicly available data might still limit the ubiquity of deep \glspl*{CNN} for human detection in both optical and \gls*{TI} aerial imagery.
Available data in the field either only consists of optical~\cite{stanford,uav123}, or only \gls*{TI} images~\cite{eth_tir,otcvbs1,ptb_tir}, and no publicly available dataset provides real-world data collected in the field for an expressive evaluation of human detection algorithms in search and rescue scenarios.
Our publication provides such a collection of optical-\gls*{TI} field datasets for extensive training and testing of human detection algorithms.
%
%
%
Furthermore, besides optimizing the raw detections, we propose a complete framework to detect, track, localize, and re-identify humans, as shown in Fig.~\ref{fig:overview}.

\section{Related Work}
\label{sec:Related Work}

Early work in the field of human detection from \glspl*{UAV}~\cite{Rudol2008,Gaszczak2011,Flynn2013,Blondel2014,Vempati2015,Kummerle2016} strongly resembled the one from object detection and applied classical methods such as Haar-like features introduced by~\cite{Viola2001}, the Felzenszwalb detector~\cite{Felzenszwalb2010}, or \gls*{HOG} features together with linear \gls*{SVM} classifiers as introduced by~\cite{Dalal2005}.
A lot of this early work~\cite{Rudol2008,Gaszczak2011,Flynn2013,Blondel2014_2,Kummerle2016} concluded that the combination of \gls*{TI} and optical images is highly beneficial for the task of human detection from high aerial views.
%
%

With the rise of deep learning, the application of \glspl*{CNN} started to become well-established for the task of human detection from aerial views~\cite{DeOliveira2016,Sathyan2016,Herrmann2016,Bejiga2017}.
By comparing against more classical feature extractors and detectors such as Haar, \gls*{HOG} or \gls*{SVM}, an improvement in detection performance as well as a better generalization when using \glspl*{CNN}, was stated throughout. 
Research in object detection progressed quickly and gradually yielded deeper, better, and faster-performing object detection networks. 
On the side of two-stage detectors, the R-CNN network and its successors ~\cite{Girshick2014,Girshick2015,Ren2015}, as well as \glspl*{FPN} by ~\cite{Lin2017_FPN} gained a lot of attention.
On the other hand, first one-stage detectors such as YOLO and its successors ~\cite{Redmon2016,Redmon2017,Redmon2018} or the \gls*{SSD} by ~\cite{Liu2016}, impressed with high frame-rates while still achieving competitive detection performance.
%
%
Some recent work in the field already started to include these state-of-the-art object detectors such as adaptations of the R-CNN network~\cite{Liu2018,Perera2018} or one-stage detectors such as YOLO9000 or \gls*{SSD}~\cite{Chang2018,Yun2019} into their pipelines.
The use of either one-stage or two-stage object detectors, however, results in a speed-accuracy trade-off, as shown by~\cite{Bondi2018}. 
This is in accordance with recent findings by~\cite{Lin2017_RetinaNet}.
Their one-stage object detector framework called RetinaNet tries to address this discrepancy between inference speed and detection performance and close the gap between state-of-the-art one-stage and two-stage object detection frameworks. Wang et al.~\cite{Wang2018}, for instance, used RetinaNet~\cite{Lin2017_RetinaNet} as a proposed solution for object detection in aerial views.
RetinaNet was evaluated against other one-stage and two-stage detectors, namely \gls*{SSD} and Faster R-CNN on the Stanford drone dataset~\cite{stanford}.
The evaluation resulted in a similar statement to the one by~\cite{Lin2017_RetinaNet}, showing state-of-the-art performance of RetinaNet compared to two-stage detectors, while running at speeds comparable to the ones obtained from one-stage detectors.
This motivates the use of these state-of-the-art one stage detectors also for our task, since their fast inference speeds are crucial when running on-board a \gls*{UAV} with limited computing power.
%

The vast majority of deep learning architectures is trained and evaluated on optical imagery.
In fact, to the best of our knowledge, the only work to date which uses a \gls*{DL}-based human detector on thermal imagery from a \gls*{UAV} was conducted by~\cite{Bondi2018} in 2018, where a Faster R-CNN framework was applied to detect poachers in thermal images.
More so, none of the publications make use of \emph{both} the optical and thermal domain together with recent state-of-the-art deep object detection networks.
Likely, this is due to the lack of publicly available data: Publicly available datasets either only consist of optical images \cite{stanford,uav123} or only thermal images \cite{eth_tir,otcvbs1,ptb_tir}.
In this paper, we approach these research gaps as follows:
Firstly, we present a comprehensive performance comparison of state-of-the-art human detection architectures, trained with optical and thermal imagery. 
In particular, the evaluation comprises YOLO and RetinaNet with different ResNet backbones. 
For reference, we additionally provide the results from a hand-crafted thermal-optical human detection pipeline \cite{Kummerle2016}.
Secondly, to the best of our knowledge, this paper constitutes the first publication on deep learning-based human detection from \glspl*{UAV} which combines optical-thermal imagery.
The employed merging strategy is explained in detail in Sec.~\ref{sec:merging_strategy}.
Thirdly, the evaluation is conducted on datasets recorded during \gls*{SAR} field tests.
%
%
The collection of annotated datasets and the implemented annotation tool is released along with this paper.
%
Despite the great success of recent findings in the field of object detection, very small sample sizes as well as strong-view point variations in aerial images, make the task challenging.
Recent work like the one by~\cite{Liu2018} or~\cite{Chang2018}, clearly shows that adaptations to current object detection networks are crucial for a decent performance on aerial images from \glspl*{UAV}. 
In this paper, we propose to use optimized custom anchors and up-scaled images to boost the detection from high altitudes and reveal this performance gain in a detailed evaluation.
Finally, we describe in detail our approach to track, localize, and re-identify humans to improve the overall performance beyond the raw detection.
%
\section{Human Detection}
\label{sec:detection}
\subsection{State-of-the-Art Object Detectors Revisited}
\label{subsec:object_detectors}

Our proposed pipeline uses a customized RetinaNet network with a ResNet50 backbone as introduced by \cite{Lin2017_RetinaNet} as a state-of-the-art one-stage object detector.
As already outlined, this is mainly due to limited computing resources on-board of \glspl*{UAV}, as well as similar detection performance of RetinaNet compared to two-stage detectors in recent publications \cite{Lin2017_RetinaNet,Wang2018}.
Our implementation follows the original implementation and introduces crucial customizations, as outlined in Sec \ref{subsec:network_customization}.
This proposed framework is then compared against YOLOv3 using a darknet-53 backbone, another state of the art one-stage object detector. 
Both networks follow a similar basic architecture collecting features at different scales using their respective backbone networks and subsequently regressing and classifying the output bounding boxes.
%
%
As one of their main contributions, RetinaNet introduces a novel focal loss, based on the well-known cross-entropy loss.
Since YOLOv3 still relies on the standard cross-entropy loss, the focal loss further constitutes the main difference between these two object detection frameworks. 
In \cite{Lin2017_RetinaNet}, Lin et al.\ state the significant imbalance of image background and foreground in many object detection tasks as the main reason for lacking the performance of conventional one-stage detectors.
They were able to show that a conventional cross-entropy loss is easily overwhelmed by the vast amount of background samples seen during training, when running regression and classification directly on top of a dense feature map.
By adding a modulating factor $\left(1 - p_t\right)^{\gamma}$ to the cross-entropy loss, they define the novel focal loss using a tunable focusing parameter $\gamma > 0$, able to counteract this large imbalance. 
%
%
\subsection{Network Customization}
\label{subsec:network_customization}

\subsubsection{Optimal Anchor Selection}
Both RetinaNet as well as YOLOv3 use nine anchor bounding boxes for final bounding box regression. 
While RetinaNet uses fixed hand-picked anchors, YOLOv3 uses k-means dimension clustering on the COCO dataset~\cite{coco} to find optimal anchors.
Furthermore, during training, the selection strategies, deciding on which anchor bounding boxes are assigned to the ground-truth bounding boxes, differ quite significantly between the two networks. 
Lin et al.~\cite{Lin2017_RetinaNet} use an adjusted assignment rule originating the region proposal network of Faster R-CNN~\cite{Ren2015}.
Anchors are assigned to ground-truth boxes for an \gls*{IOU} value of 0.5 or higher and to background for \gls*{IOU} values in $\left[0, 0.4\right)$.
Each anchor is assigned to at most one ground-truth box, and all remaining unassigned anchors are ignored during training.
Redmon et al.~\cite{Redmon2018}, on the other hand, do not use such a dual \gls*{IOU} threshold.
Instead, they simply assign one anchor per ground-truth bounding box by using the anchor with the largest \gls*{IOU} value.
If an anchor is not the best but overlaps a ground-truth box with more than 0.5 \gls*{IOU}, the prediction is ignored.
All other anchors, not assigned to any ground-truth boxes, do only incur a loss for the object prediction.
The need for customized anchors in our pipeline is crucial:
Using the standard anchors for RetinaNet on our training dataset showed that a lot of samples in both optical and thermal images did not contribute to the training process because of their really small size and not reaching an \gls*{IOU} value of $0.4$ or higher with any of the standard anchor bounding boxes.  
Resolving this by changing the added anchor sizes at each level of the original publication by \cite{Lin2017_RetinaNet} to custom sizes of $\left\lbrace 2^{-2}, 2^{-1}, 2^{0} \right\rbrace$, the number of bounding boxes contributing to the training was sufficiently increased.
More specifically, the amount of optical samples contributing to the final stage of training was increased from 57.7\% to 97\% on the optical side and from 33.2\% to 88.1\% on the thermal side of our training dataset.



\subsubsection{Image Resolution Augmentation}
A second natural way of improving the detection of really small bounding boxes is to simply augment their sizes.
Final detection performance can be improved significantly, by either using augmented- or higher-resolution images, as also stated by Bejiga et al.~\cite{Bejiga2017}.
This can be achieved by increasing the input image size during inference since both networks being fully-convolutional.
Doubling the standard image sizes of the RetinaNet variants was able to bring a considerable performance improvement as shown in Sec.~\ref{sec:Results}.

\subsection{Optical-Infrared Merging Strategy}
\label{sec:merging_strategy}
\begin{figure}[htb]
 \begin{subfigure}[b]{0.48\linewidth}
    \includegraphics[width = \textwidth]{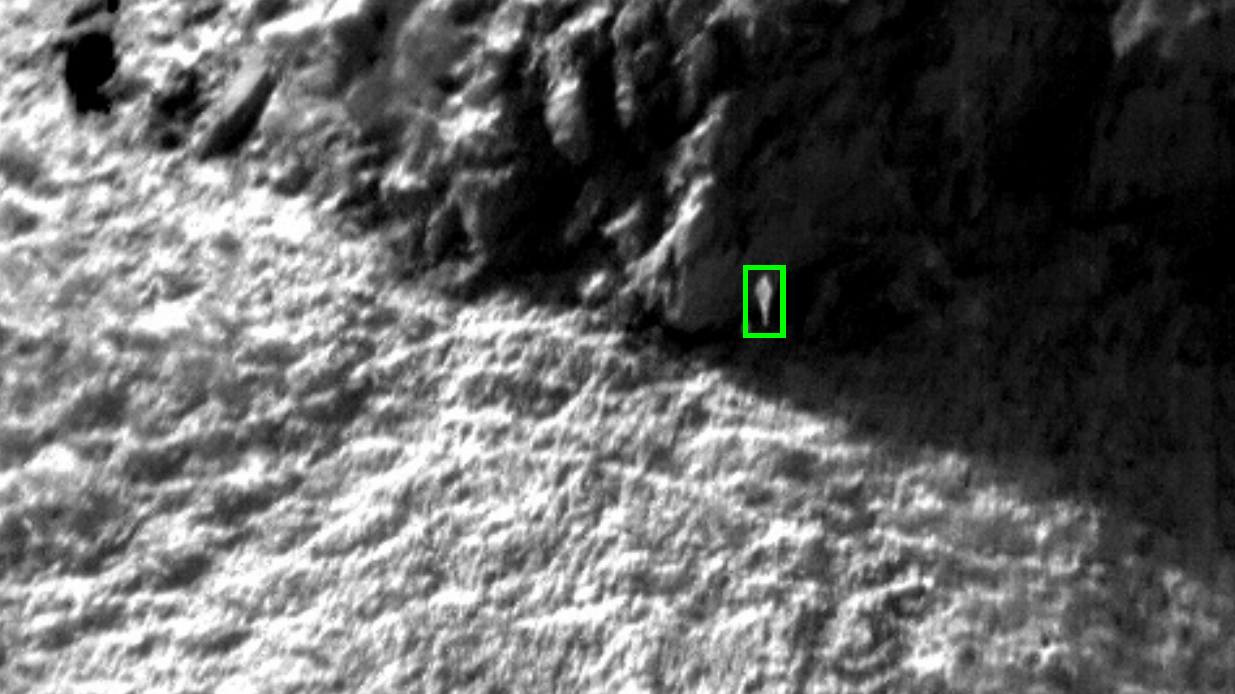}
    \caption{}
  \end{subfigure}
  \hfill
  \begin{subfigure}[b]{0.48\linewidth}
    \includegraphics[width = \textwidth]{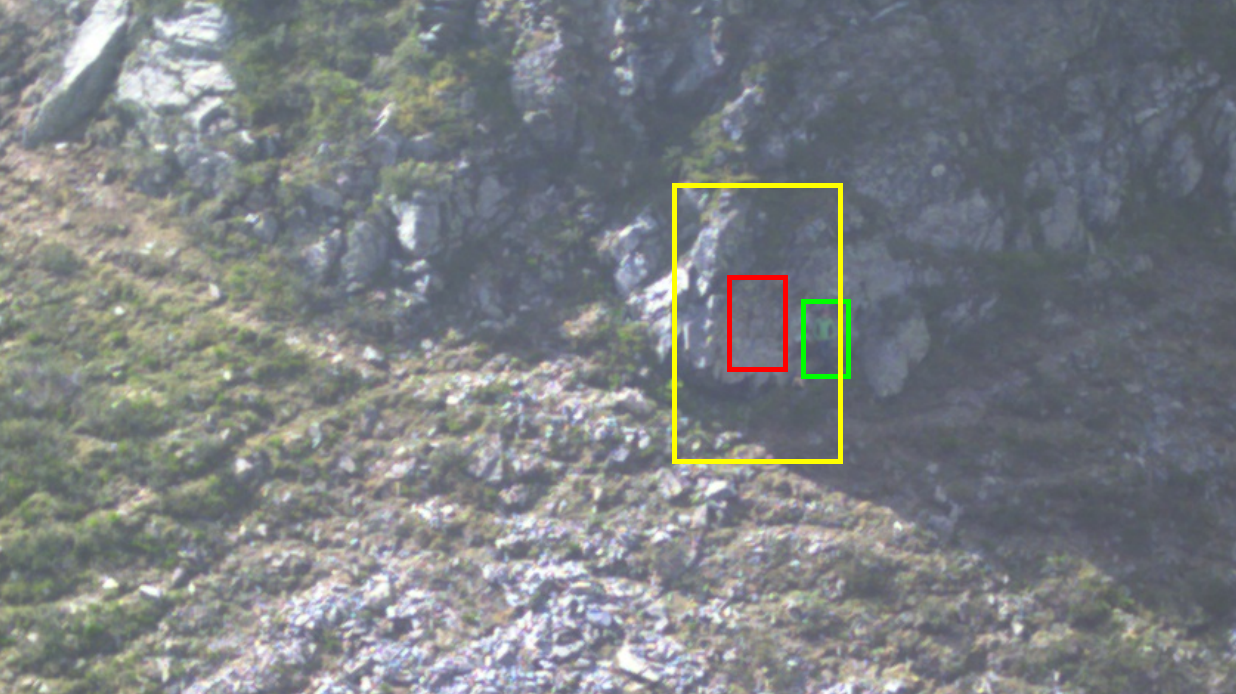}
        \caption{}
  \end{subfigure}
  \caption{Exemplary matching process. Original detection in the thermal image on the left in green. Raw mapped bounding box in the right optical image in red, sliding window area in yellow and final matched bounding box (detection) in green.}
  \label{pics:sliding_window_example}
\end{figure}

The proposed pipeline utilizes optical and thermal imagery.
This is especially useful to further reduce false positives during day flights, for detecting humans during night flights, or to find humans in aggravated optical conditions, for instance, when obscured by shadows due to large rocks or trees. 
%
%
The camera intrinsic and extrinsic parameters are used to map feature positions from the optical to the thermal image and vice versa.
However, small inaccuracies in the calibration, the image triggering, or exposure time may result in pixel offsets in both spectra.
Especially in applications with fast-moving and fast-turning \glspl*{UAV} this may become an issue. 
To account for these inaccuracies, we propose to use a sliding window to match bounding boxes between the spectra, as illustrated in Fig.~\ref{pics:sliding_window_example}.
More specifically, a rectangle nine times the size of the mapped bounding box is searched for a target bounding box using 36 sliding window steps in total.
Matching of the sliding window and target bounding boxes is done similarly to the standard procedure proposed by \cite{Dollar2009}, using an \gls*{IOU} threshold of 0.5. 
After a completed matching step, a logical OR merging scenario is used, considering all bounding boxes from both domains while averaging the respective prediction scores.

\subsection{Network Training}
To train the network, publicly available, external datasets, as well as internal data, recorded by one of our \glspl*{UAV}, are used.
Using the implemented annotation tool, a total of $11$ optical sequences consisting of $\numprint{40445}$ images, and $14$ thermal sequences containing $\numprint{1029}$ images are available for training.
The human bounding boxes in theses sequences are annotated with a distinct ID for each individual human, attributes for the pose (upright, sitting or lying), and an occlusion attribute.
Altogether, the newly collected datasets add up to a total of $\numprint{34022}$ human bounding boxes in the optical images, and $\numprint{37228}$ human bounding boxes in the thermal images.
%
%
In addition to our internal datasets, all available public datasets containing humans from both optical \cite{minidrone, stanford, uav123} and thermal \cite{eth_tir,otcvbs1,otcvbs11,ptb_tir,vot_tir} aerial views have been gathered.
Together with these external datasets, the final amount of available data at hand consisted of around $\numprint{1000000}$ annotations in over $\numprint{170000}$ optical images, and around $\numprint{200000}$ annotations in over $\numprint{100000}$ thermal images. 


The final training of the two object detectors was conducted using a two-stage training procedure:
Starting with pre-trained Imagenet \cite{Russakovsky2015} weights, an initial pre-training step was carried out using all the data at hand, including our newly collected dataset and all available external datasets.
The best results in pre-training were achieved by freezing all the backbone weights for both RetinaNet and YOLOv3 and only training and adapting all other, randomly initialized weights to the novel domain. Subsequently, the resulting weights of the pre-training step were then used to initialize a final fine-tuning step.
In this step, only the newly collected, domain-specific data was used to train all the layers of both the networks. 
Training of the two object detectors was carried out according to the original publications of RetinaNet \cite{Lin2017_RetinaNet} and YOLOv3 \cite{Redmon2018}.
Both networks were trained on a cluster using Nvidia GeForce GTX 1080 Ti GPUs.
While YOLOv3 originally uses a smaller sized input image together with a multi-scale training strategy \cite{Redmon2018}, RetinaNet uses a larger single size input image during training.
To be able to train both networks using a similar batch size of eight images, RetinaNet was trained on a total of eight GPUs while YOLOv3 was trainable on a single GPU. Both networks include data augmentation and other training features mentioned in the original publications.
Randomly selected sequences out of the external datasets were used as validation sets in pre-training and a fixed sequence of our own dataset was used as validation set for the final fine-tuning training scenario. 

%
\section{Human Localization and Re-Detection}
This section describes our approach to track, localize, and re-detect humans in the optical spectrum based on qualitative results. \newline

\begin{figure}[h!]
\centering
\begin{minipage}[t]{1\linewidth}
\centering
\includegraphics[width=0.45\linewidth,keepaspectratio]{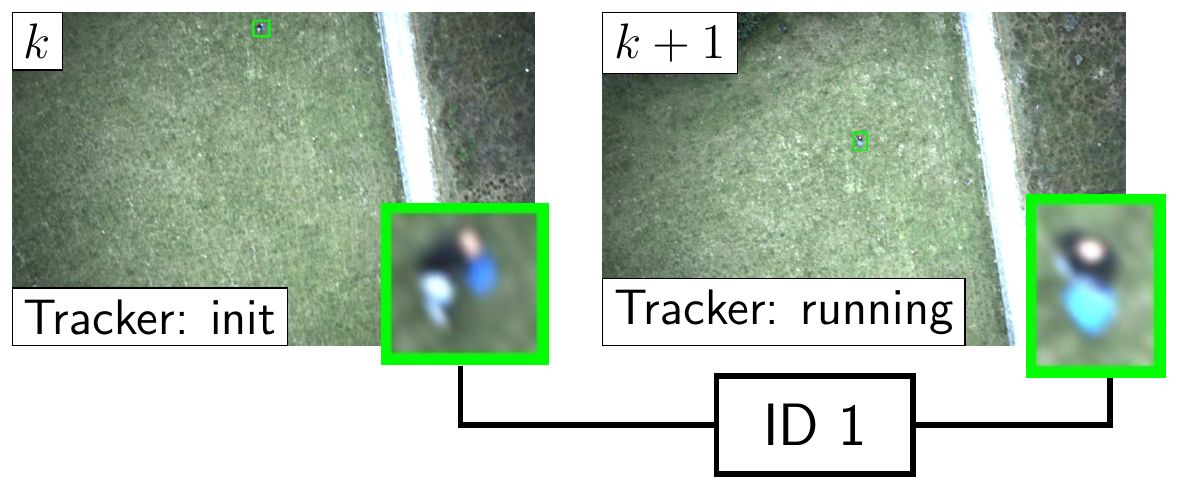}
\captionof{figure}{Human tracking: Humans that are tracked across subsequent frames ($k$, $k+1$) are assigned the same human ID.}
\label{fig:diagram_1}
\end{minipage}
\end{figure}

\begin{figure}[h!]
\centering
\begin{minipage}[t]{1\linewidth}
\centering
\includegraphics[width=0.65\linewidth,keepaspectratio]{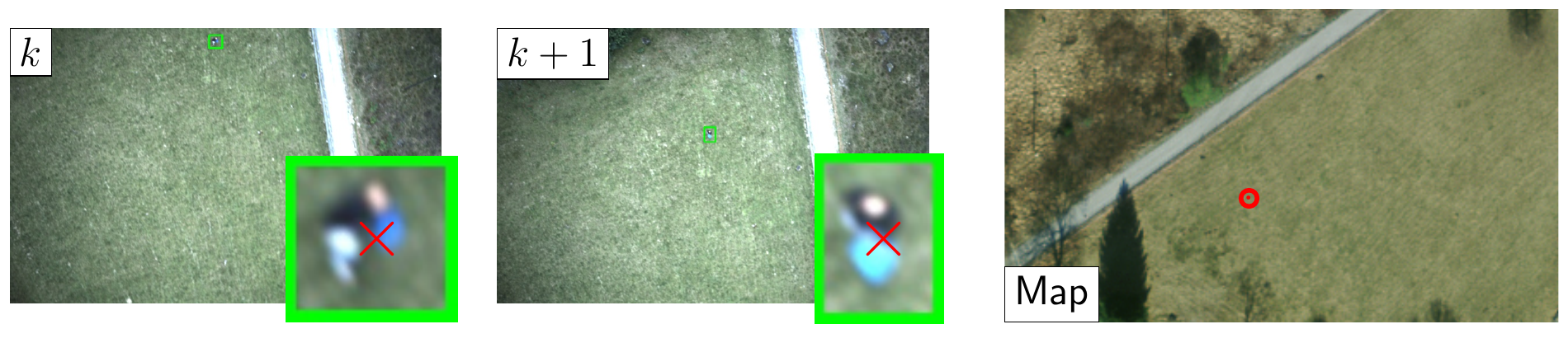}
\captionof{figure}{Human localization: Based on two subsequent detections, the geo-referenced human position can be triangulated, as visualized in the satellite orthoimage.}
\label{fig:diagram_2}
\end{minipage}
\end{figure}

\begin{figure}[h!]
\centering
\begin{minipage}[t]{1\linewidth}
\centering
\includegraphics[width=0.6\linewidth,keepaspectratio]{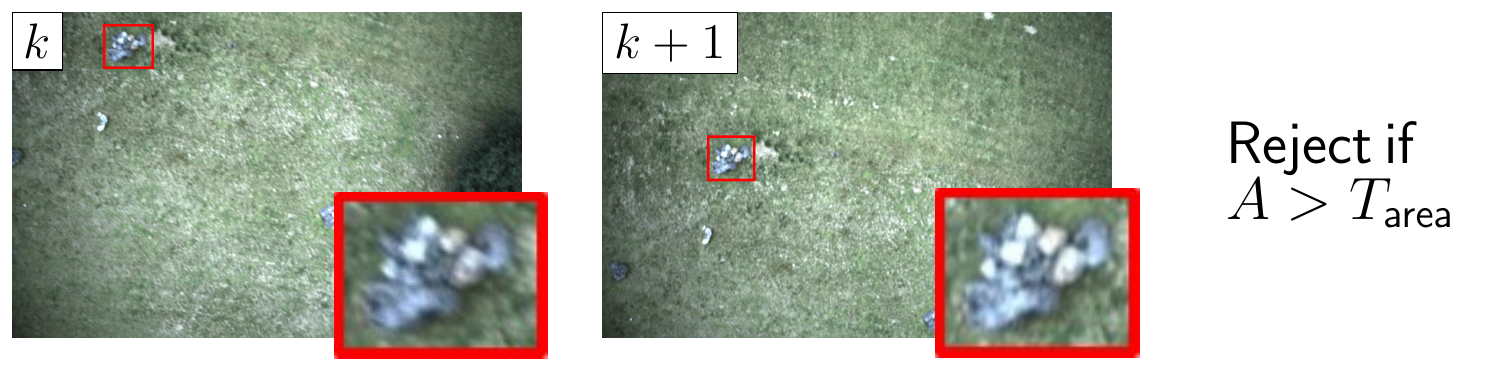}
\captionof{figure}{Metric outlier rejection: The detection is rejected if the estimated metric area $A$ of the bounding box is above a threshold $T_\text{area}$.}
\label{fig:diagram_3}
\end{minipage}
\end{figure}

\begin{figure}[htb]
\centering
\begin{minipage}[t]{\linewidth}
\centering
\includegraphics[width=0.9\linewidth,keepaspectratio]{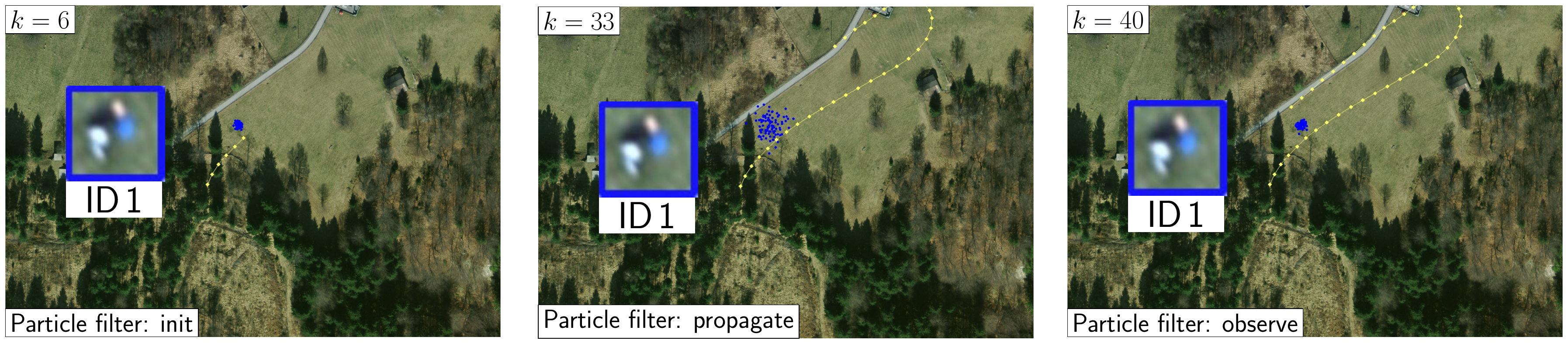}
\captionof{figure}{Particle filter: The particles for human ID 1 are visualized as blue points, the trajectory flown by the Rega drone is shown in yellow.}
\label{fig:diagram_4}
\end{minipage}
\end{figure}

\begin{figure}[htb]
\begin{minipage}[t]{0.5\linewidth}
\centering
\includegraphics[width=0.7\linewidth]{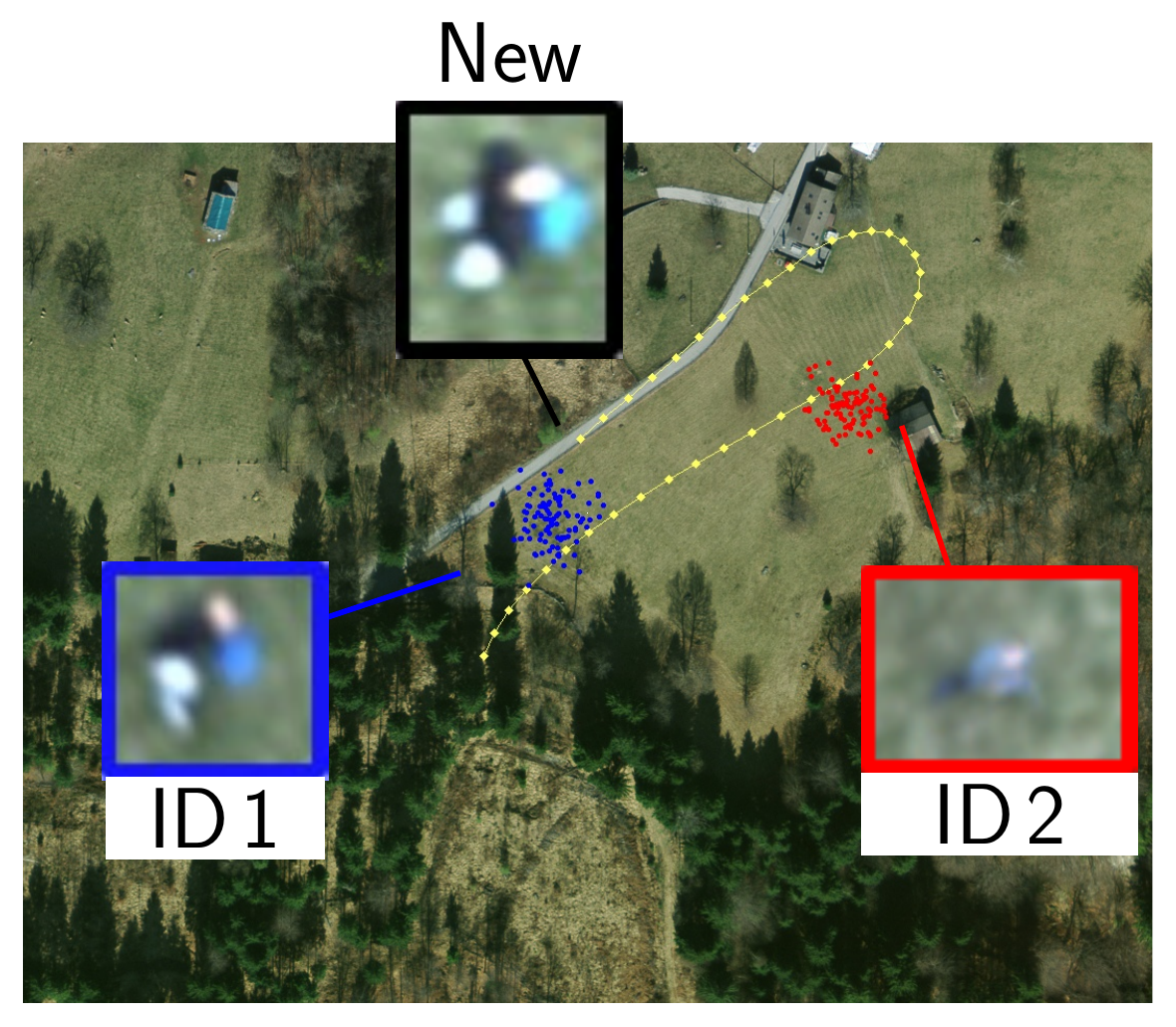}
\end{minipage}
\hfill
\begin{minipage}[t]{0.45\linewidth}
\includegraphics[width=0.8\linewidth]{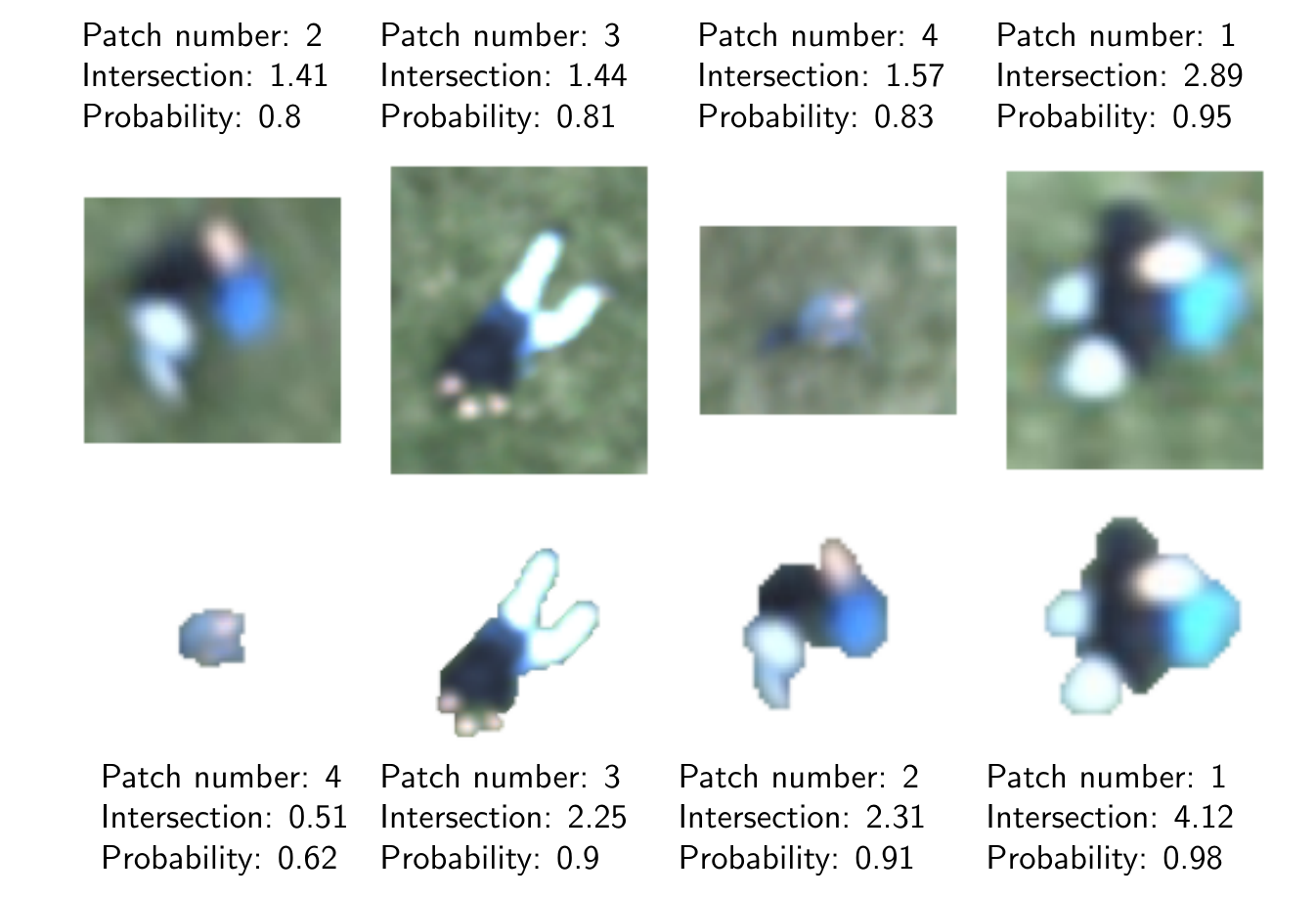}
\end{minipage}
\caption{Human re-detection: The algorithm needs to decide if the new detection can be associated to an already observed human or if it is, in fact, a new victim. On the right, the histogram similarity is evaluated based on the Intersection metric. The query image is patch number 1.}
\label{fig:diagram_5}
\end{figure}

\paragraph{Human Tracking:} For every observation classified as human the detector described in Sec.~\ref{sec:detection} creates a new victim ID.
However, the final goal of the proposed framework is to associate every victim with a unique ID and to compute its position or path in 3D.
In a first step towards this goal, an object tracker bundles all detections of a victim as long as the human is within the field of view of the camera (cf.\ Fig.~\ref{fig:diagram_1}).
The human tracking is tested on the optical image stream with a frame-rate of $\unit[4]{Hz}$, resulting in potentially large pixel displacements of an observation between two subsequent frames. 
To reduce the pixel displacement and simultaneously increase the speed of the tracker the image is half-sampled.
Object trackers implemented in \cite{bradski2000opencv} were tested, including MIL \cite{Babenko09}, KCF \cite{henriques2012exploiting}, GOTURN \cite{held2016learning}, among which CSRT \cite{lukezic2017discriminative} performed best and was selected.
\paragraph{Human Localization:} Given a track of observations in the form of bounding boxes and the corresponding camera poses, the 3D path of the object can be estimated.
The camera poses are assumed to be given as input to the framework.
As the human may be non-static, two-view triangulation of consecutive observations are used.
The center of the bounding box is selected for triangulation, as illustrated in Fig.~\ref{fig:diagram_2}.
\paragraph{Metric Outlier Rejection: }
The metric bounding box area is used to reject false detections as follows:
Given two consecutive observations of an object and the triangulated 3D object position ${\mathbf{p}^W_{h}}$, the depth $d$ can be inferred via ${d = \|\mathbf{p}^W_\text{UAV} -\mathbf{p}^W_{h} \|_2}$.
The depth is then used to transfer the bounding box from pixel coordinates to meters, resulting in the points $\mathbf{p}^W_{i}, i=(1,...,4)$ (clockwise).
The metric area of the bounding box is then ${ A = 0.5 (\overline{\mathbf{p}^W_1 \mathbf{p}^W_2} \times \overline{\mathbf{p}^W_1 \mathbf{p}^W_3} + \overline{\mathbf{p}^W_1 \mathbf{p}^W_3}\times \overline{\mathbf{p}^W_1 \mathbf{p}^W_4})}$.
Objects with bounding box areas $A>T_\text{area}$ are classified as outliers and rejected as visualized in Fig.~\ref{fig:diagram_3}.
%
%
%
\paragraph{Particle Filter: } 
To handle occlusions, re-detections, and to incorporate a probabilistic motion model, a \gls*{PF} is initialized for every human.
For computational reasons the \gls*{PF} is restricted to two dimensions $x$ and $y$ in \gls*{UTM} coordinates.
If necessary, the altitude of the victim can be queried from the existing map.
The \gls*{PF} implementation and notation closely follows \cite{Simon2006} (cf.\ Fig.~\ref{fig:diagram_4}):
\begin{itemize}[noitemsep]
\item \textbf{Initialization:} Given the first triangulated position, denoted as $\mathbf{z}_0=\left(x,y\right)$  with $\sigma_{\mathbf{z}}=3$, randomly draw $N=100$ initial particles $\mathbf{x}^{+}_{0,i}$ with $i = \{1,...,N\}$ from $\mathcal{N}\sim(\mu=\mathbf{z}_0, \sigma=\sigma_{\mathbf{z}})$.
\item \textbf{Propagation: } In the propagation step, compute the a priori particles $\mathbf{x}^{-}_{k,i}$ given a random walk motion model, assuming a maximum human velocity $\mathbf{v}_\text{max} = \unit[1.2]{\frac{m}{s}}$ in both, the $x$- and $y$-direction:\\
${\mathbf{x}^{-}_{k,i} = f_{k-1}(\mathbf{x}^{+}_{k-1,i}, \mathbf{w}^i_{k-1})} = {\mathbf{x}^{+}_{k-1,i} + \mathbf{w}^i_{k-1}\Delta t}$ where ${\mathbf{w}^i_{k-1}}$ is a random velocity $\smash{\mathbf{v}_\text{rand}}$ drawn from the uniform distribution $\mathcal{U}\sim(-\smash{v_\text{max}}, {v_\text{max}})$ and $\smash{\Delta t}$ is the time difference between two consecutive frames in seconds.
\item \textbf{Measurement:} Given a new observation $\mathbf{z}_k$ associated with the victim ID, compute the relative likelihood $q_i$ of this observation for every particle $\mathbf{x}^{-}_{k,i}$ by evaluating the conditional \gls*{PDF} $p(\mathbf{z}_k|\mathbf{x}^{-}_{k,i})$ based on the measurement equation: \newline ${q_i \sim \alpha^{-1} \exp(-0.5 (\mathbf{z}_k-\mathbf{x}^{-}_{k,i})^\top \mathbf{R}^{-1} (\mathbf{z}_k-\mathbf{x}^{-}_{k,i}))}$ with $\alpha=2\pi \det(\mathbf{R})^{\frac{1}{2}}$ and measurement noise $\mathbf{R}=\sigma_{\mathbf{z}} \cdot \mathbf{I}_{2\times 2}$.
Normalize via ${q_i = \beta^{-1} q_i }$ with ${\beta=\sum_{j=1}^N q_j}$.
\item \textbf{Resampling:} Draw posteriori particles ${\mathbf{x}^{+}_{k,i}}$ based on normalized likelihoods $\smash{q_i}$ using \gls*{SR}~\cite{Kozierski2013}.
\end{itemize}
\paragraph{Human re-detection: } If the \gls*{UAV} flies over a previously visited area and the detector returns an object classified as human, the algorithm needs to decide if the new detection $z$ can be associated to an already observed human $h_i$ or if it is, in fact, a new victim (cf.\ Fig.~\ref{fig:diagram_5}).

Applying the Bayes theorem the probability $p(h_i|z)$ that the new observation $z$ belongs to human $h_i$ can be computed with $p(h_i|z) =\gamma^{-1} p(z|h_i)p(h_i)$, $\gamma = \sum p(z|h_i) p(h_i)$.
Appearance-based and spatial information are used to compute $p(h_i|z)$.
Firstly, the conditional probability $p(z|h_i)$ is computed based on the triangulated location of the new observation and all existing humans currently tracked by \glspl*{PF}.
If $p(z|h_i)p(h_i) < T_\text{redetect} \forall i$ a new victim with corresponding \gls*{PF} is initialized.
Otherwise the new detection is associated with $\arg\max_{h_i} p(h_i|z)$, i.e., with that human that maximizes the detection probability.

Secondly, the prior $p(h_i)$ which is the probability of observing human $h_i$ is inferred from the similarity between a patch from human $h_i$ and the newly detected human patch using a binary classifier.
Since the humans are detected from a high distance we base our decision if two patches contain the same person solely on the color information.
Tab.~\ref{tab:hist_no_subtraction} presents the similarity between patches computed by comparing their color histograms using the metrics \cite{Bradski2000} Correlation, Chi-square, Intersection, and Bhattacharyya \cite{bhattacharyya1943measure}.
The patch numbers 1 to 4 correspond to the detections shown in Fig.~\ref{fig:diagram_5}.
Note that for Correlation and Intersection, higher values correspond to a higher similarity.
In contrast, for the metric Chi-square and Bhattacharyya, the lower the value, the higher the similarity.
Patch numbers 1, 2, and 3 are observations of the same human.
However, this is not reflected by the results in Tab.~\ref{tab:hist_no_subtraction} as the background has an influence on the color histogram.
Therefore, using the GrabCut algorithm \cite{rother2004grabcut}, the background is automatically subtracted (cf.\ Fig.~\ref{fig:diagram_5}) before computing the histogram which improves the re-identification results, as shown in Tab.~\ref{tab:hist_subtraction}.
Using the sigmoid function, for instance, the results from the Intersection metric are mapped to values between 0 and 1, as shown in  Fig.~\ref{fig:diagram_5}.
\begin{table}[htb]
\begin{minipage}[t]{0.45\linewidth}
\centering
\resizebox{0.9\linewidth}{!}{\begin{tabular}{l|c|c|c|c|}
 & \multicolumn{4}{c}{\textbf{Patch}} \\ 
\textbf{Method} & 1 & 2 & 3 & 4 \\\hline 
Correlation & \cellcolor{green!70} 1.0 &  0.71 & \cellcolor{green!50} 0.83 & \cellcolor{green!50} 0.83 \\\hline
Chi-Square & \cellcolor{green!70} 0.0 &  8.85 & \cellcolor{green!50} 2.39 & \cellcolor{green!30} 2.52\\\hline
Intersection & \cellcolor{green!70} 2.89 &   1.41 & \cellcolor{green!20} 1.44 & \cellcolor{green!50} 1.57 \\\hline
Bhattacharyya & \cellcolor{green!70} 0.0 & 0.53 & \cellcolor{green!50} 0.37 & \cellcolor{green!20} 0.44 \\\hline
\end{tabular}}
\caption{Quantitative results for Fig.~\ref{fig:diagram_5}: Patch similarity \mbox{\textbf{without}} background subtraction and histogram comparison.}
\label{tab:hist_no_subtraction}
\end{minipage}
\hfill
\begin{minipage}[t]{0.45\linewidth}
\centering
\resizebox{0.9\linewidth}{!}{\begin{tabular}{l|c|c|c|c|}
 & \multicolumn{4}{c}{\textbf{Patch}} \\ 
\textbf{Method} & 1 & 2 & 3 & 4 \\\hline
Correlation &  \cellcolor{green!70} 1.0 & \cellcolor{green!20} 0.57 &  \cellcolor{green!50} 0.83 & 0.09\\\hline
Chi-Square &  \cellcolor{green!70} 0.0 &  \cellcolor{green!20} 12.33 &  \cellcolor{green!50} 1.9 &  44.92 \\\hline
Intersection & \cellcolor{green!70} 4.12 &  \cellcolor{green!50} 2.31 & \cellcolor{green!20} 2.25 &  0.51\\\hline
Bhattacharyya &  \cellcolor{green!70} 0.0 &  \cellcolor{green!20} 0.52 &  \cellcolor{green!50} 0.41 &   0.83 \\\hline
\end{tabular}}
\caption{Quantitative results for Fig.~\ref{fig:diagram_5}: Patch similarity \textbf{with} background subtraction and histogram comparison.}
\label{tab:hist_subtraction}
\end{minipage}
\end{table}

\section{Hardware \& Experiment Preparation}

\subsection{Platform and Sensors}

\begin{figure}[b]
\begin{subfigure}[b]{.25\textwidth}
\includegraphics[width=\textwidth, trim=100 0 0 900, clip]{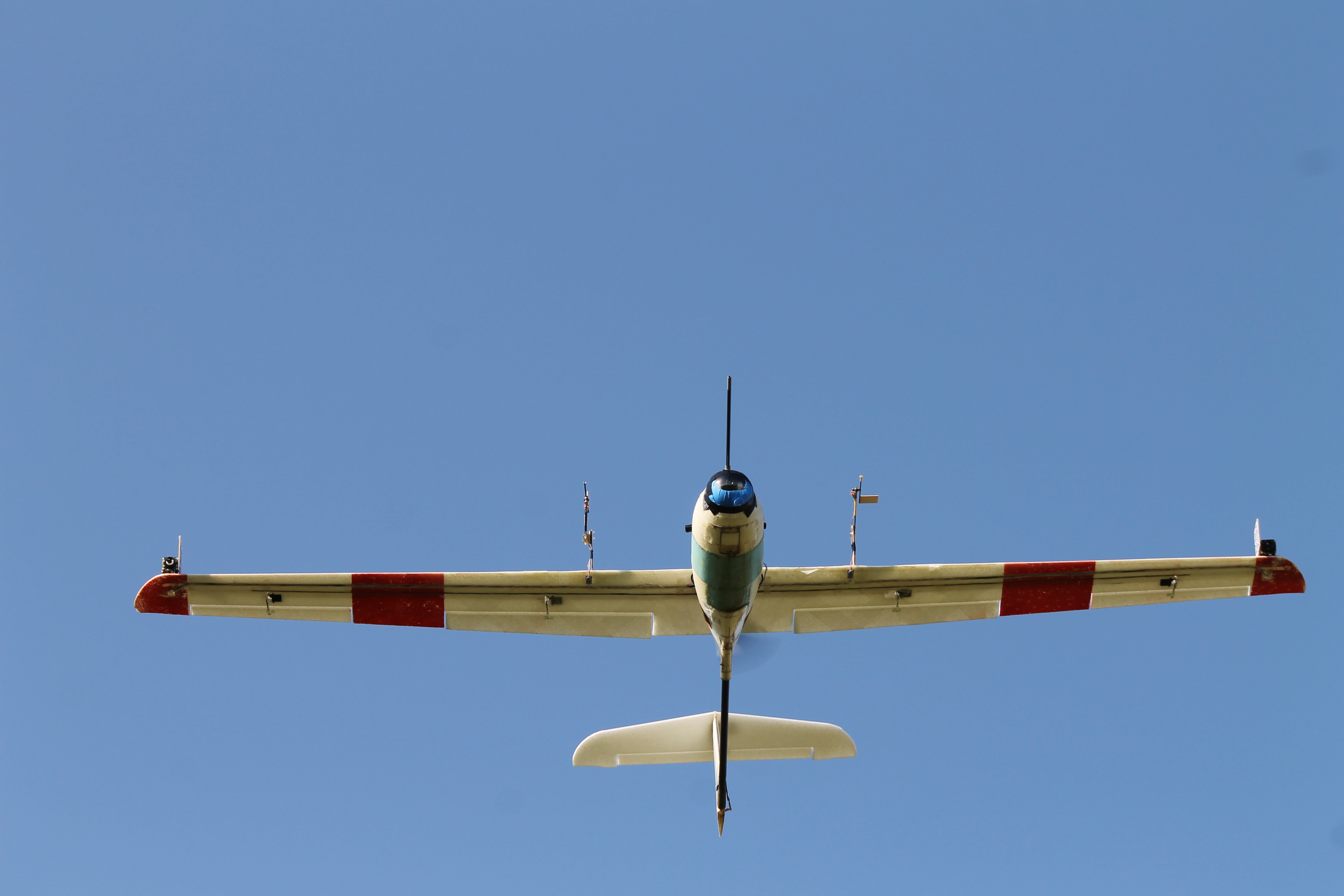}
\caption{Techpod}\label{fig:techpod}
\vspace{2ex}
\includegraphics[width=\textwidth]{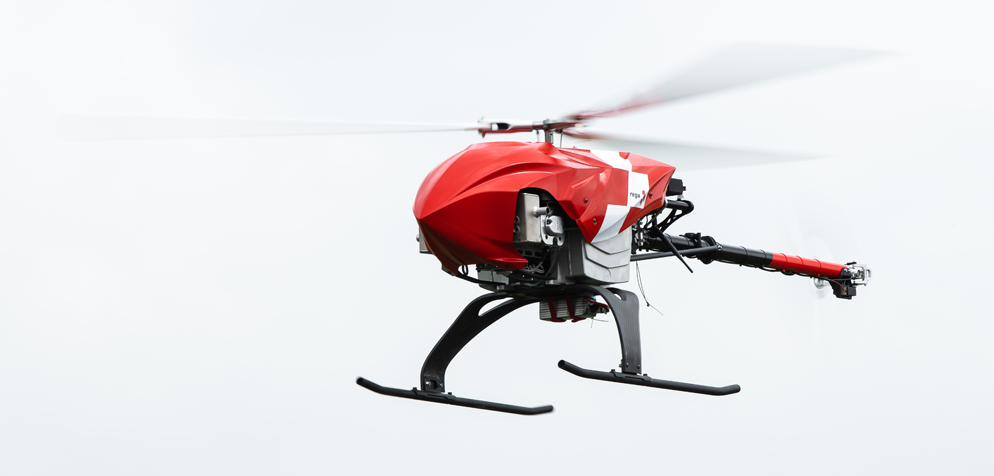}
\caption{Rega drone}\label{fig:rega_drone}
\end{subfigure}\qquad
\centering
\begin{subfigure}[b]{.55\textwidth}
\centering
\includegraphics[width=0.7\textwidth]{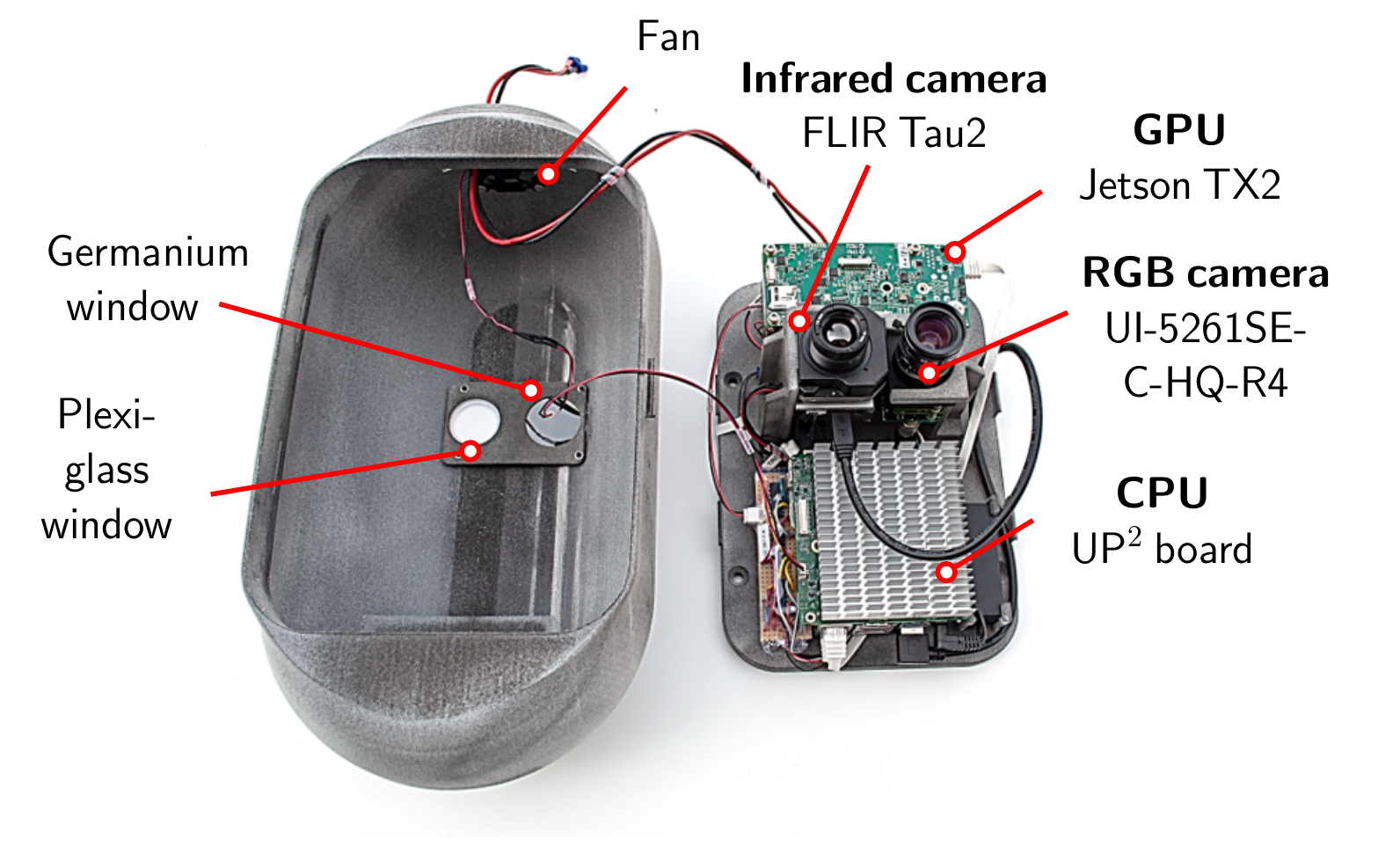}
\caption{Sensorpod with optical-infrared stereo rig.}
\label{fig:sensorpod}
\end{subfigure}
\caption{The sensorpod with optical-infrared stereo rig carried by the Rega drone.}
\end{figure}

The sensorpod used for our experiments is shown in Fig.~\ref{fig:sensorpod}.
It consists of an IDS UI-5261SE-C-HQ-R4 $\unit[1.92]{MP}$ RGB camera with a C-mount lens and a horizontal field of view of $\unit[32]{^{\circ}}$ and an infrared camera FLIR Tau2 with a resolution of $640\times 480$ pixel.
The FLIR Tau2 is mounted on a Teax image grabber with USB2 interface.
The GPU is a Jetson TX2 mounted on an Auvidea J120 carrier board and is used for network inference at run-time.
It is connected to the CPU, an UP${}^2$ board with Intel Atom Processor, via Ethernet.
The CPU has a MSATA storage of $\unit[1]{TB}$ and handles the triggering of RGB camera, infrared camera, and ADIS16448 \gls*{IMU}.
The casing features a plexiglass and Germanium window for the optical and infrared camera, respectively, and a fan for additional air circulation.
The two \gls*{UAV} platforms that are used for the various test flights are shown in Fig.~\ref{fig:techpod} and \ref{fig:rega_drone}.
\subsection{Geometric Optical-Infrared Camera and Camera-IMU Calibration}
The camera intrinsics (focal length, principal point), distortion parameters, and optical-infrared extrinsics (relative pose between the cameras and \gls*{IMU}) are required for optical-infrared image fusion and metric human localization.
For this purpose, different optical-infrared, i.e., dual-modal calibration targets have been developed and improved over time.
A dual-modal calibration target allows to directly calibrate the relative pose between the thermal and optical camera without the need for a camera-\gls*{IMU} calibration as an intermediate step.
That is the optical-infrared camera can be calibrated as a stereo rig with one single calibration dataset using Kalibr \cite{Rehder2012}.
The evolution of our developed calibration targets is represented by Tab.~\ref{tab:calibration_target_own} and was inspired by the related publications listed in Tab.~\ref{tab:calibration_targets}. 
The different calibration targets shown in Tab.~\ref{tab:calibration_target_own} and \ref{tab:calibration_targets}  are classified based on the taxonomy proposed in \cite{Rangel2014}:
Initial works utilized classical optical checkerboard calibration targets and heated the target with a flood lamp (cf. target \ref{target1} or \cite{Prakash2006}).
Based on the emissivity difference of black and white squares, the pattern can be made visible also in the \gls*{TI} spectrum.
However, fuzzy transitions between black and white edges lead to missed or inaccurate corner detections and unsatisfying calibration results.
Based on this finding, we attempted to increase the sharpness of the edges by using  materials with contrary emissivity properties.
For this we designed target \ref{target2} which is made out of black colored wood (emissivity $\approx 0.8$ \cite{Lide2004}) and aluminum (emissivity $\approx 0.095$ \cite{Lide2004}).
Likewise, the authors of \cite{Skala2011, Vidas2012, Saponaro2015} focused on improving the contrast of checkerboard targets using different materials and masks. 
However, as for instance pointed out by \cite{Yu2013}, the corner detection remains inherently error-prone in the \gls*{TI} spectrum, and instead, the usage of circular features is advised.
Our final dual-modal calibration target \ref{target3} consists of $6\times 7$ circular features, laser-cut into an aluminum calibration target, and filled with machine-cut black wooden plates.
\begin{table}[htb]
\centering 
\resizebox{\textwidth}{!}{\begin{tabular}{|l|p{2.3cm}|p{2.3cm}|c|c|l|l|l|}\hline
& \textbf{Picture IR} & \textbf{Picture RGB} & \textbf{RGB} & \textbf{IR} & \makecell[c]{\textbf{Target type}} & \textbf{Working Principle}  &  \textbf{Features} \\ \hline
 \rotatebox{90}{\newtag{No.1}{target1}} & \makecell[c]{\includegraphics[width=\linewidth]{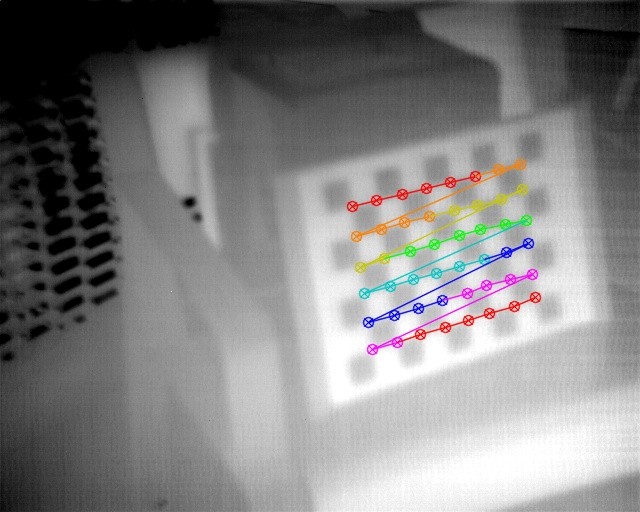}} & \makecell[c]{\includegraphics[width=\linewidth]{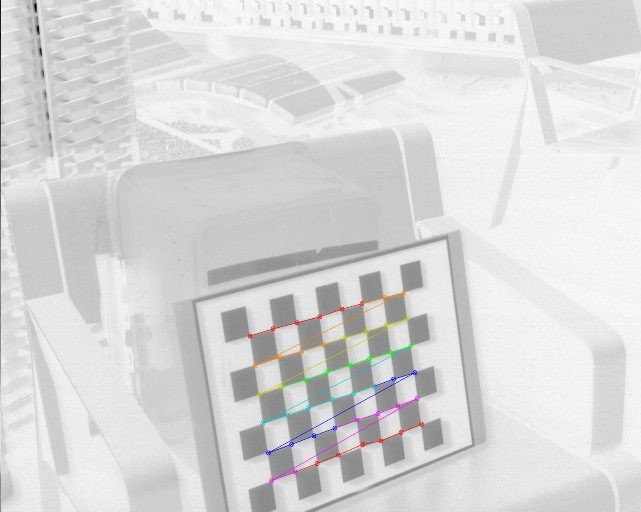}} & \checkmark & \checkmark & \makecell[l]{\textbf{Checkerboard} \\ Paper print-out \\ glued on wood} & \makecell[l]{Color emissivity  \\ difference (black, white)} & \makecell[l]{Corners \\ $6\times 8$ squares \\ $\unit[0.036]{m}$} \\ \hline
 \rotatebox{90}{\newtag{No.2}{target2}} & \makecell[c]{\includegraphics[width=\linewidth]{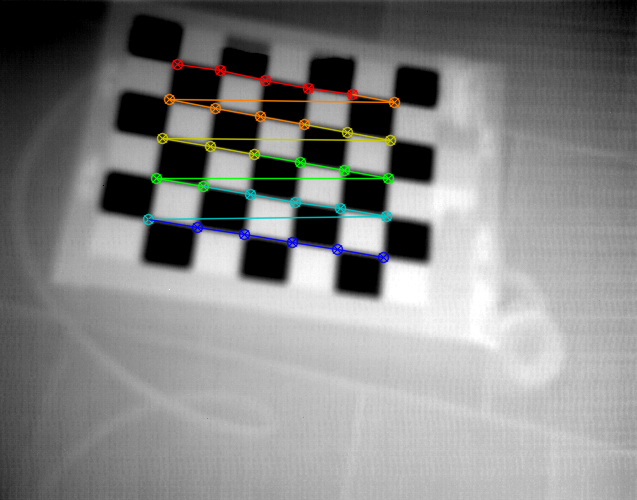}} & \makecell[c]{\includegraphics[width=\linewidth]{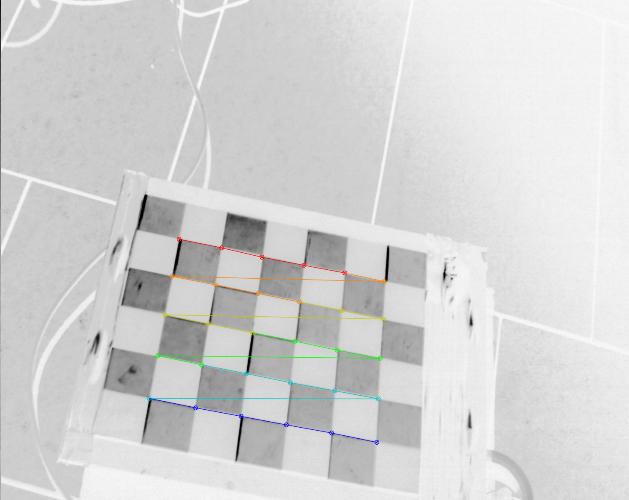}} & \checkmark & \checkmark & \makecell[l]{\textbf{Checkerboard}  \\ Black colored \\  wooden squares,\\ Aluminum squares}& \makecell[l]{Color emissivity \\ difference (black, white) \\ Material emissivity difference \\(wood, aluminum)} & \makecell[l]{Corners \\ $5\times 6$ squares \\ $\unit[0.036]{m}$}  \\ \hline
 \rotatebox{90}{\newtag{No.3}{target3}} & \makecell[c]{\includegraphics[height=1.5cm]{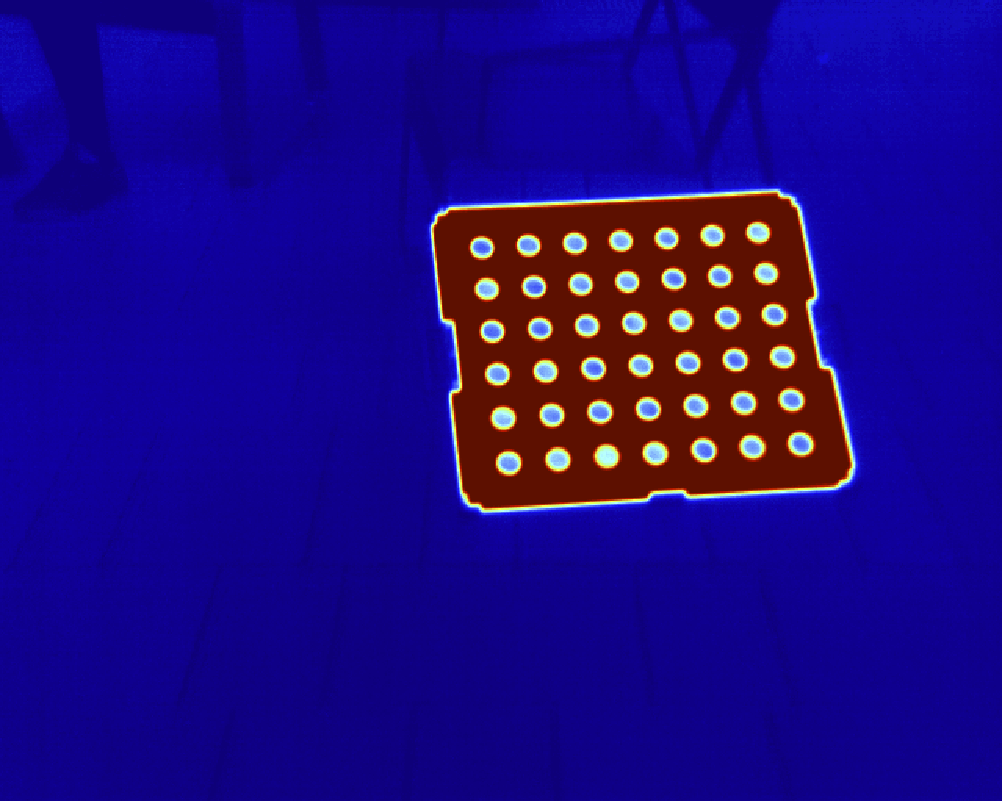}} & \makecell[c]{\includegraphics[width=\linewidth]{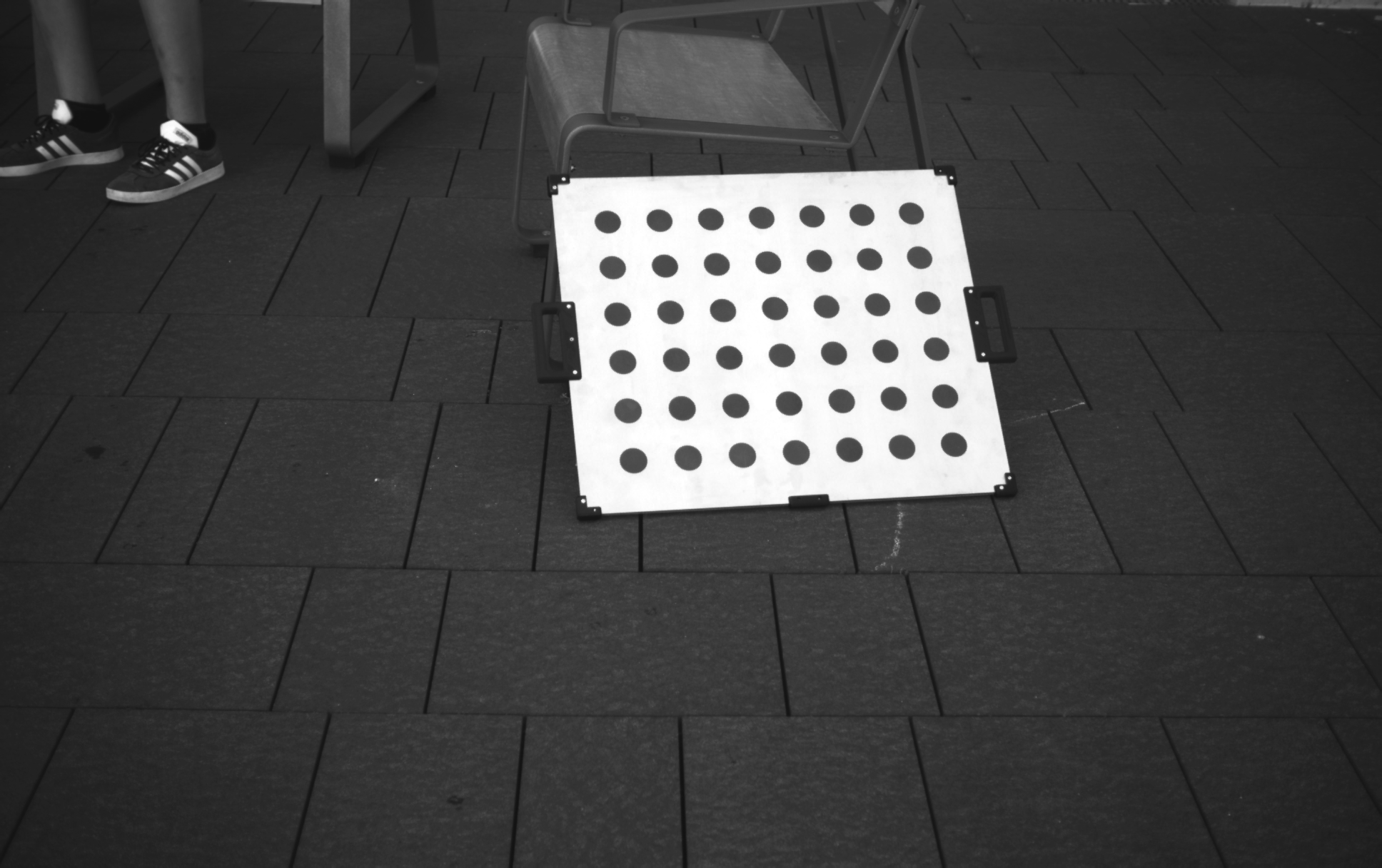}} & \checkmark & \checkmark & \makecell[l]{\textbf{Square grid} \\ Aluminum target \\ with inserted \\ black wooden \\ circles} & \makecell[l]{Color emissivity \\ difference (black, white)\\ Material emissivity difference \\ (wood, aluminum)} & \makecell[l]{Circles \\ $7\times 6$ circles\\ $\unit[0.08]{m}$}  \\ \hline
\end{tabular}}
\caption{Evolution of our dual-modal calibration targets (sorted by date).}
\label{tab:calibration_target_own}
\end{table}
Based on the detections, the camera intrinsics and extrinsics are calibrated with Kalibr \cite{Rehder2012}.
To be able to use Kalibr, the infrared images are inverted and thresholded. 
We obtained the best calibration results for datasets that are recorded outside where the calibration target is facing the clear sky.
For this target, Kalibr reports the following errors: For a single camera calibration $\sigma=0.09$ (infrared) and  $\sigma=0.09$ (optical); for a camera-\gls*{IMU} calibration $e_\text{reproj}=0.60$ (infrared) and $e_\text{reproj}=0.25$ (optical).

\begin{table}[htb]
\begin{minipage}[t]{1\textwidth}
\centering 
\resizebox{\textwidth}{!}{\begin{tabular}{|p{2cm}|c|c|c|l|l|l|}\hline
\textbf{Publication} & \textbf{Picture} & \textbf{RGB} & \textbf{IR} & \makecell[c]{\textbf{Target type}} & \textbf{Working Principle}  &  \textbf{Features} \\ \hline
\cite{Prakash2006} & \makecell[c]{\includegraphics[width=0.08\linewidth]{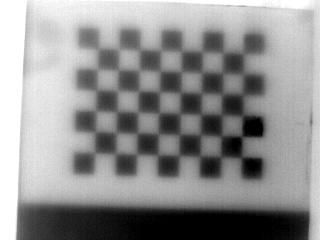}} & \makecell[c]{\checkmark \footnote{Not shown in the paper but in principle possible}} & \checkmark & Checkerboard & \makecell[l]{
Flood lamp \\Color emissivity difference (black, white)} & Corners  \\ \hline
\cite{Skala2011} & \makecell[c]{\includegraphics[width=0.08\linewidth]{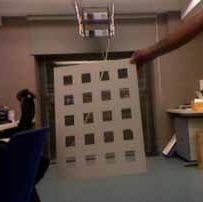}} & \checkmark & \checkmark & Hermann grid & \makecell[l]{Material emissivity difference (Styrofoam, air)}&  Corners  \\ \hline
\cite{Vidas2012} & \makecell[c]{\includegraphics[width=0.08\linewidth]{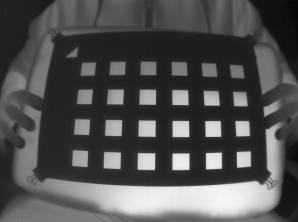}} & \checkmark & \checkmark &  Hermann grid &  \makecell[l]{Heated backdrop, e.g. monitor\\ Material emissivity difference (cardboard, monitor)\\
difference in temperature and/or\\difference in thermal emissivity} &  Corners \\ \hline
\cite{Yu2013} &\makecell[c]{\includegraphics[width=0.08\linewidth]{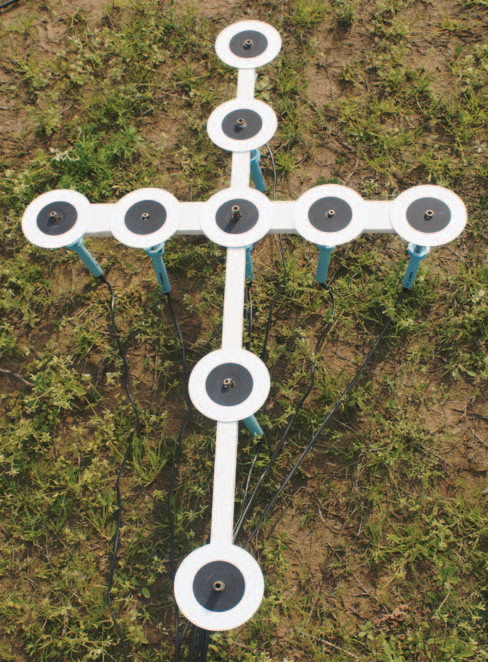}} & \checkmark  & \checkmark & Cross & \makecell[l]{Difference in temperature (Thermostatic heaters) \\
Color emissivity difference (black, white)} & Circles  \\ \hline
\cite{Saponaro2015} &\makecell[c]{\includegraphics[width=0.08\linewidth]{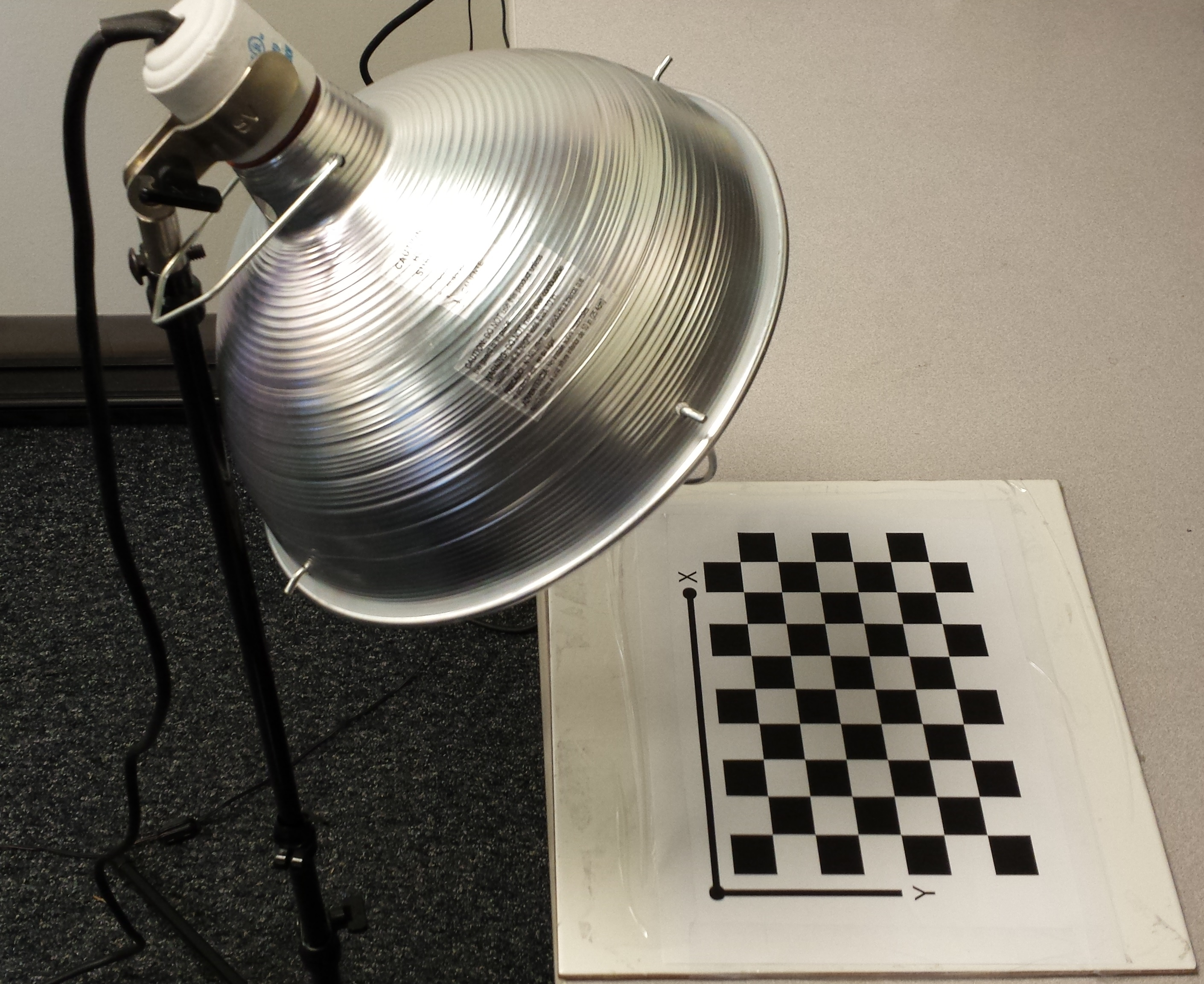}} & \checkmark  & \checkmark & Checkerboard & \makecell[l]{Flood light \\
Color emissivity difference (black, white) \\
Material emissivity difference (ceramic, paper)}  & Corners \\ \hline
\cite{Filippov2019} &\makecell[c]{\includegraphics[width=0.08\linewidth]{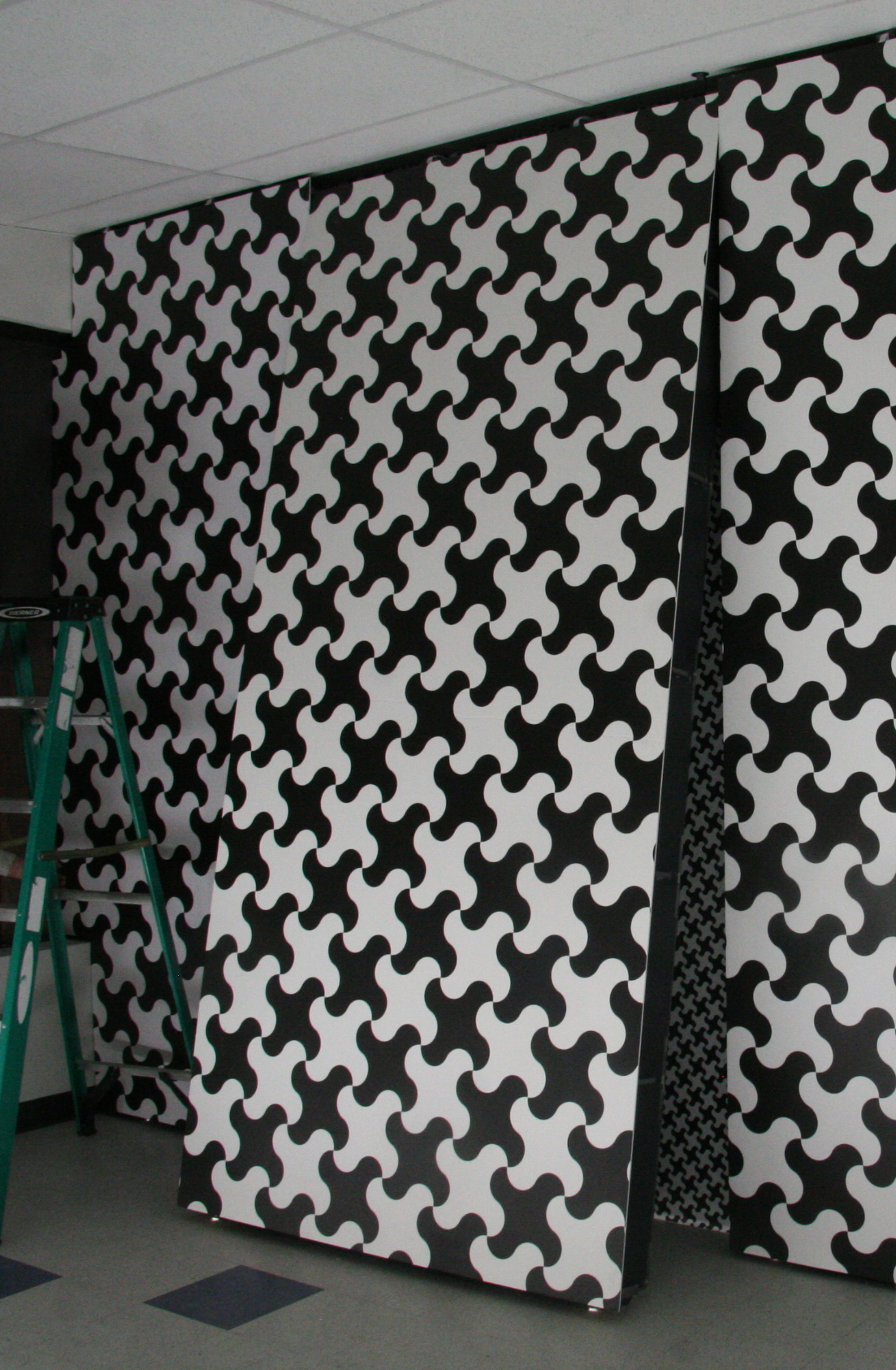}} & \checkmark & \checkmark & Wall & \makecell[l]{Temperature difference (flood lights, fan) \\ Color emissivity difference(black, white) \\ Material emissivity difference (foam, aluminum)} & Arcs  \\ \hline
\end{tabular}}
\caption{Optical and infrared calibration methods (sorted by publication date)}
\label{tab:calibration_targets}
\end{minipage}
\end{table}
%
\subsection{Optical-Infrared Dataset Collection and Annotation}
\begin{figure}[htb]
   \centering
\resizebox{0.7\textwidth}{!}{\begin{tabular}{cc}
\includegraphics[width=.4\linewidth]{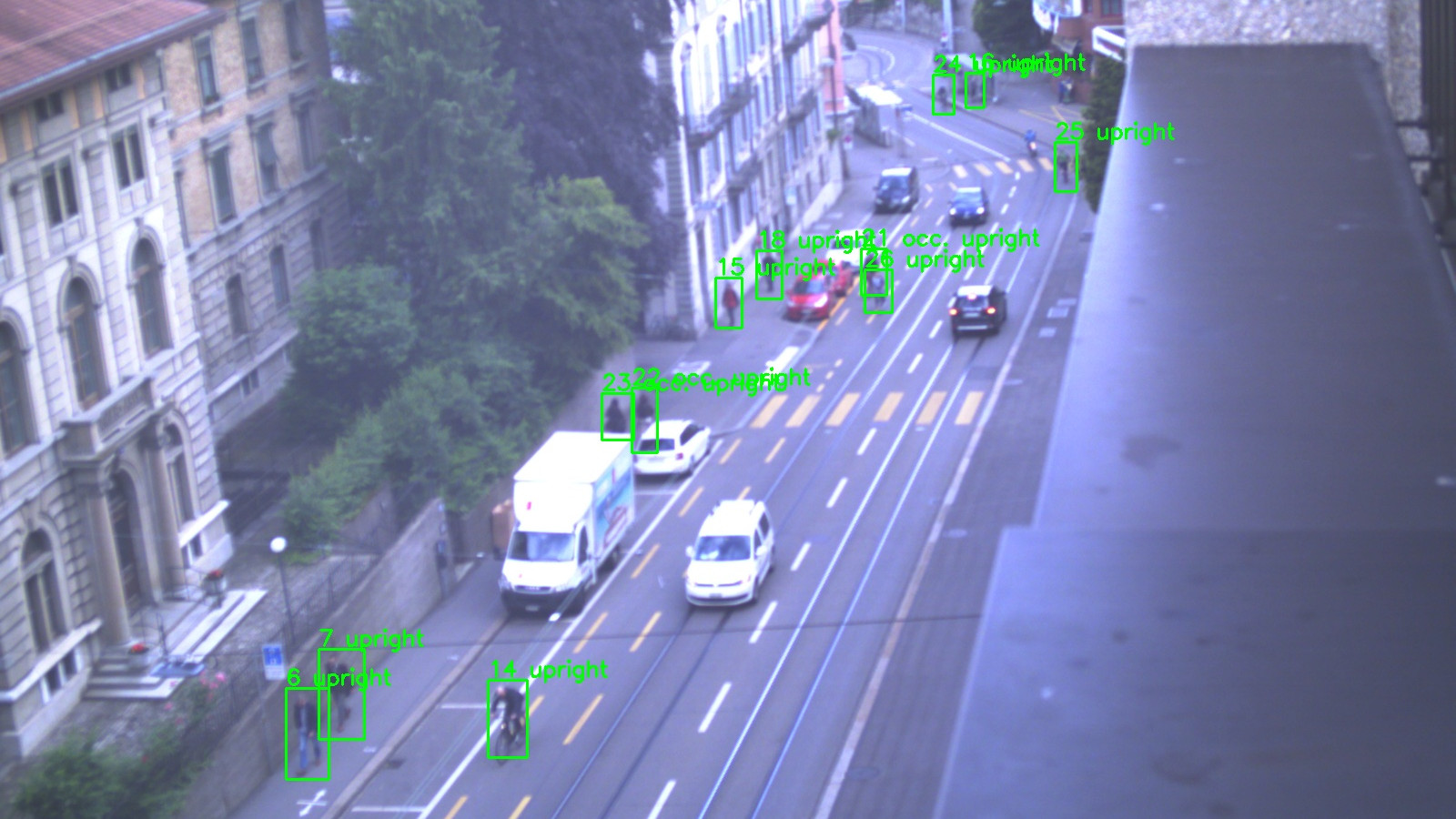} \hfill
\includegraphics[width=.4\linewidth]{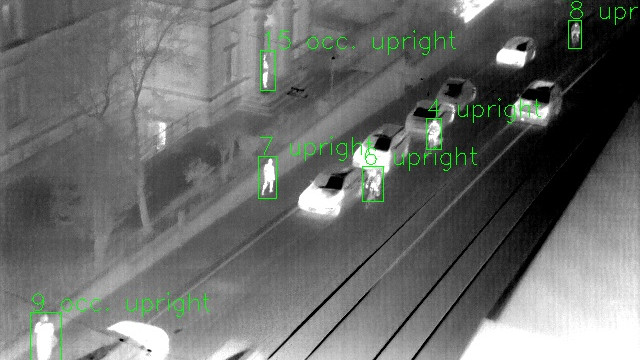} 
\\
\includegraphics[width=.4\linewidth]{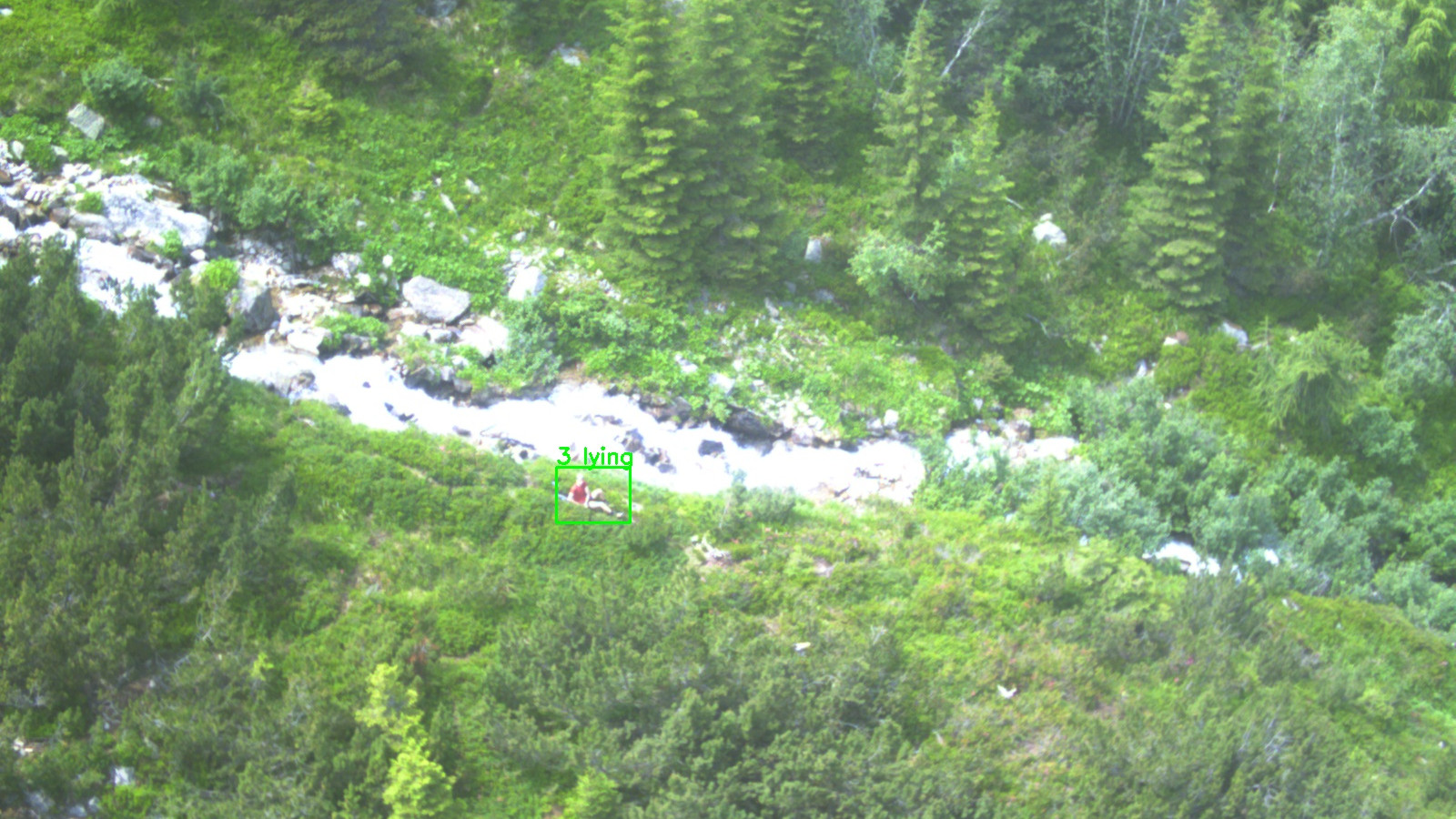} \hfill
\includegraphics[width=.4\linewidth]{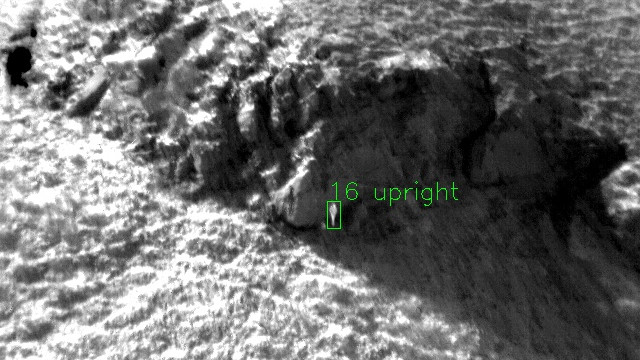}
\\
\includegraphics[width=.4\linewidth]{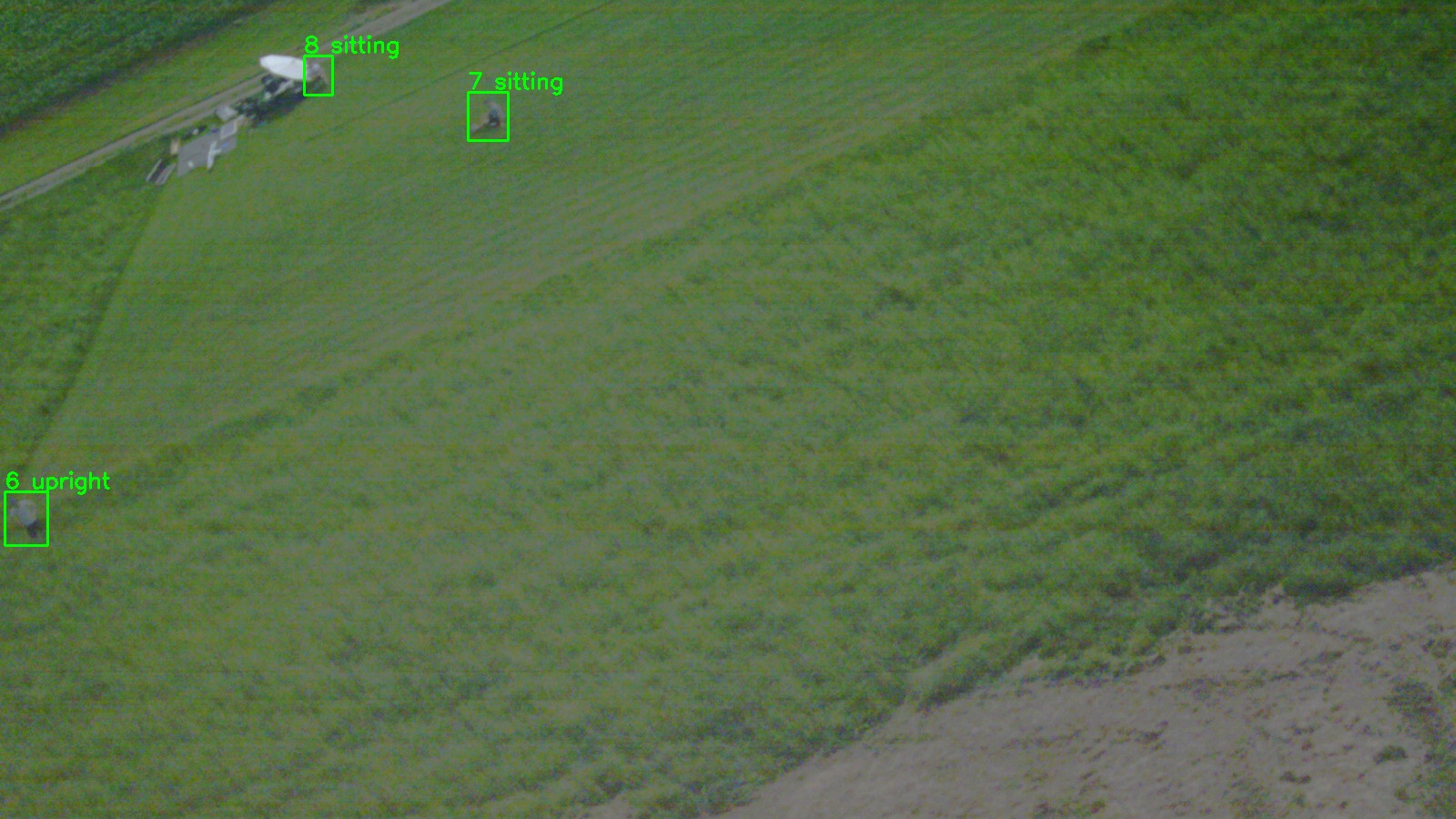} \hfill
\includegraphics[width=.4\linewidth]{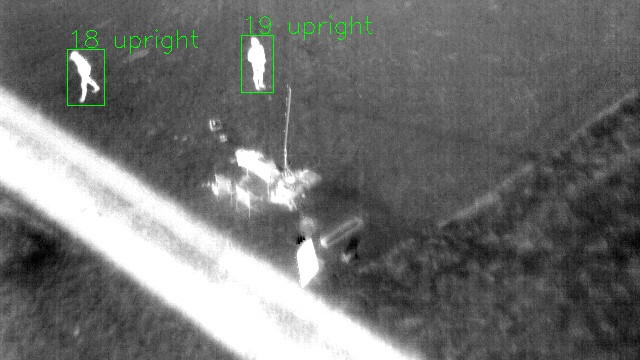} 
\\
\includegraphics[width=.4\linewidth]{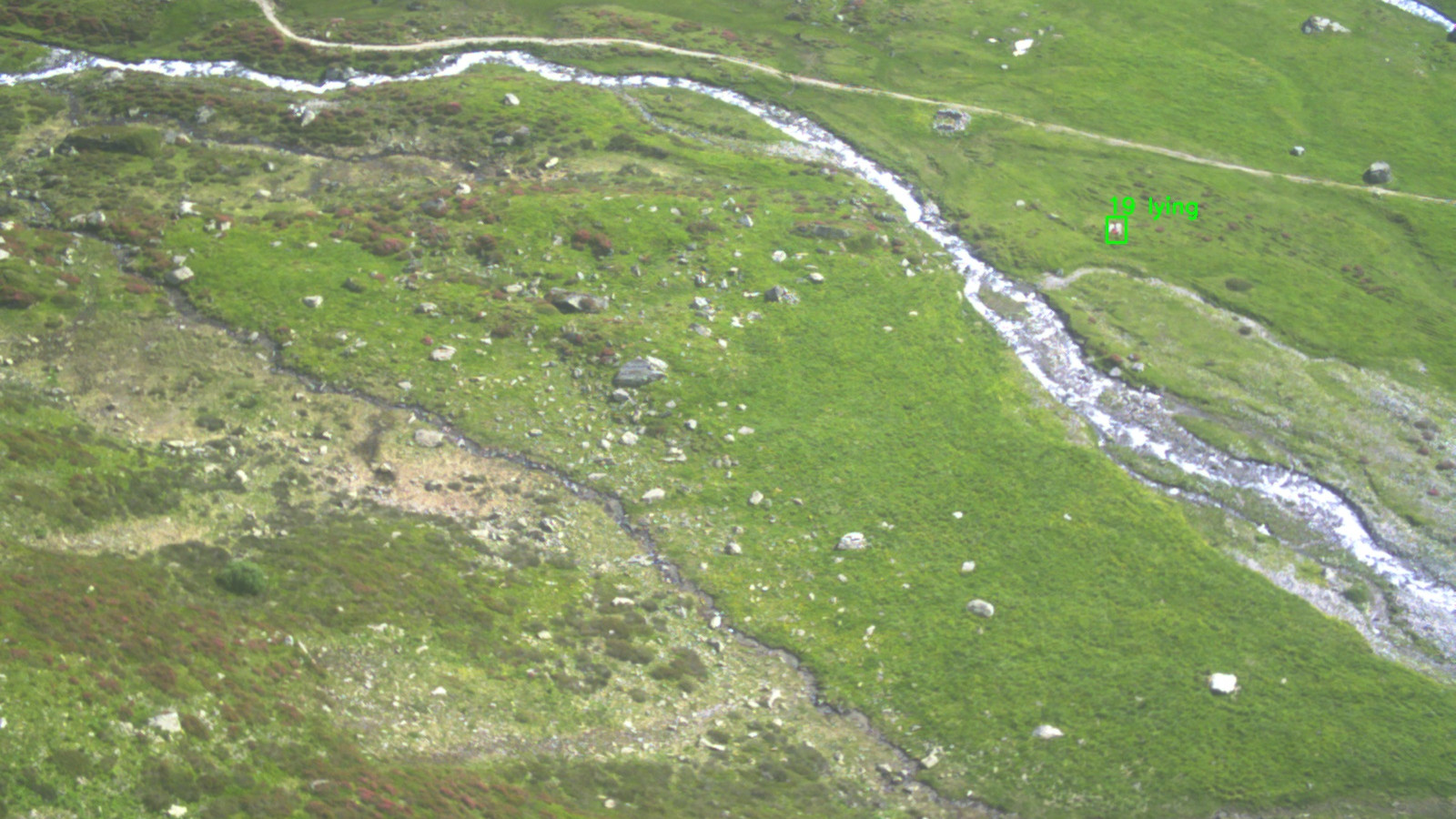} \hfill
\includegraphics[width=.4\linewidth]{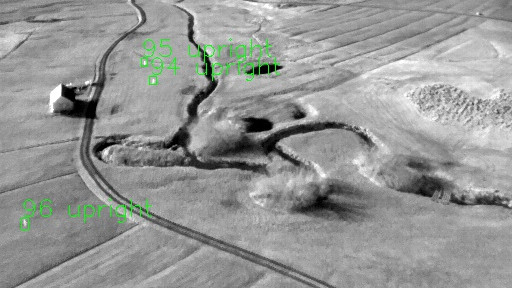}
\\
\includegraphics[width=.4\linewidth]{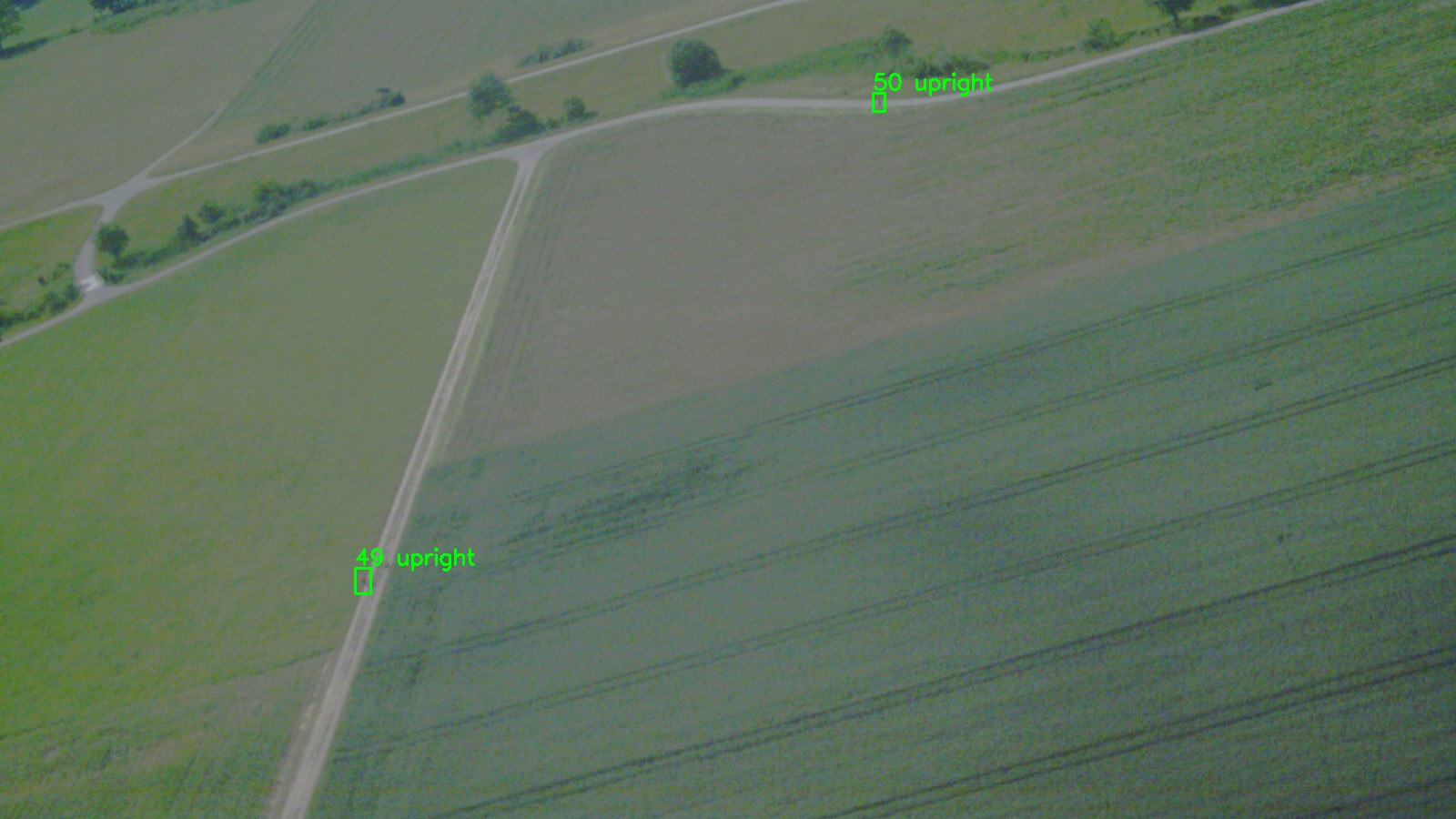} \hfill
\includegraphics[width=.4\linewidth]{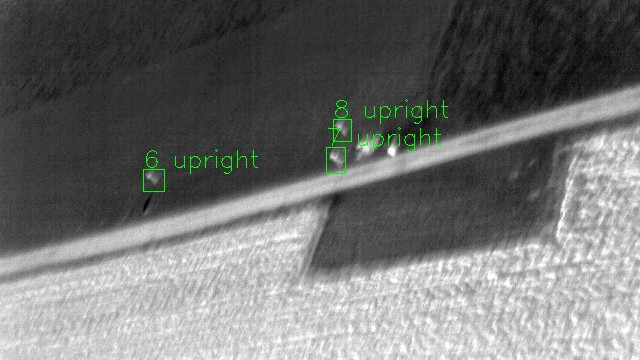} 
\\
\includegraphics[width=.4\linewidth]{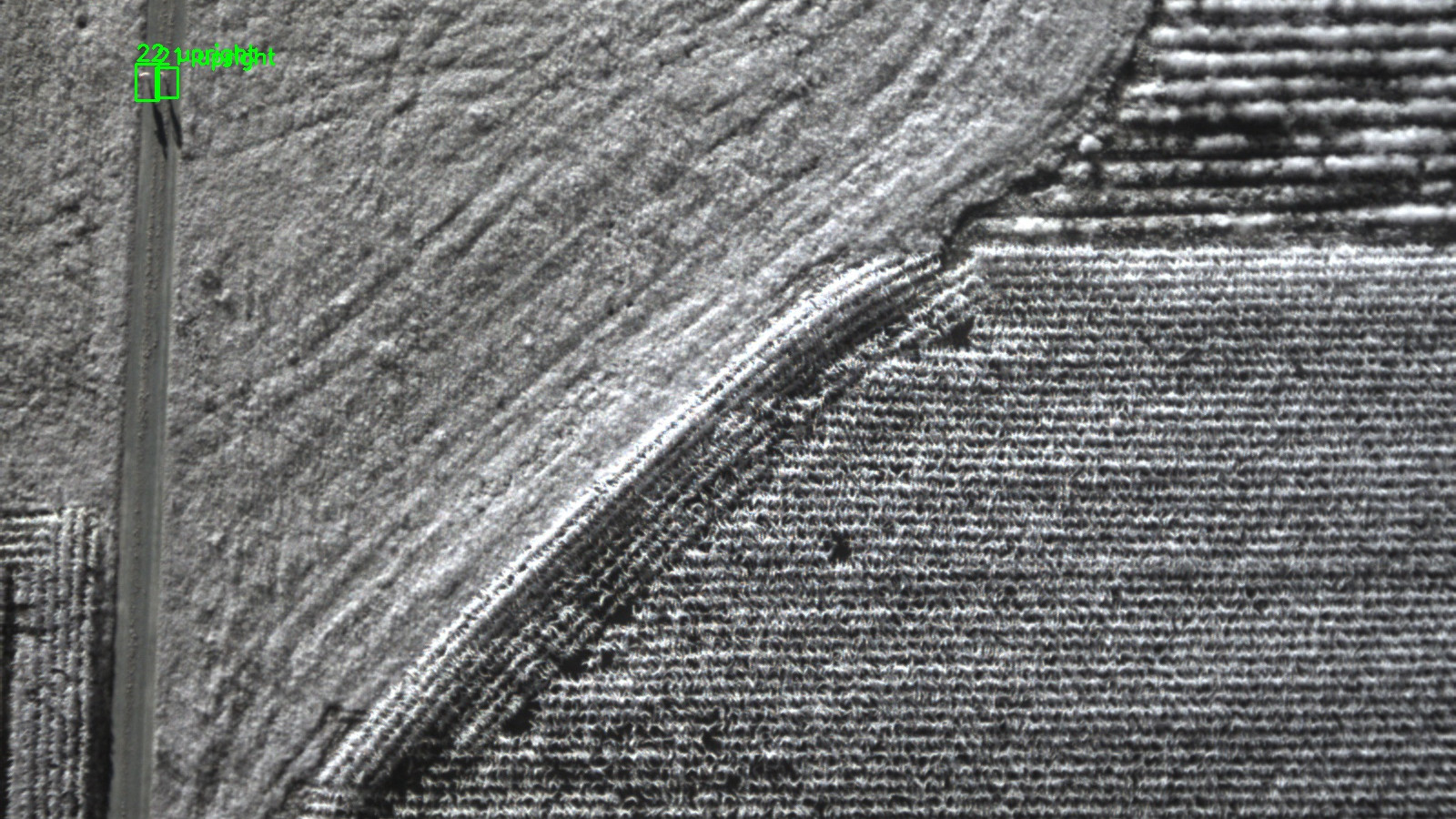} \hfill
\includegraphics[width=.4\linewidth]{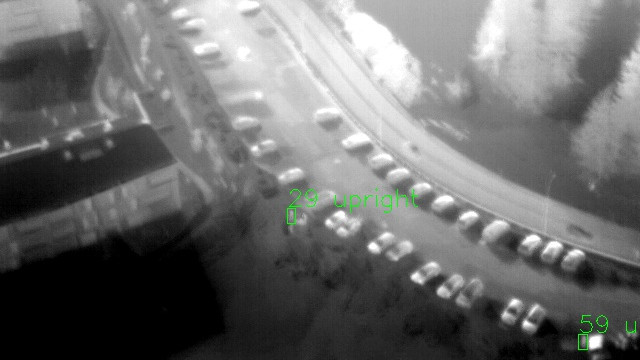}
\end{tabular}}
\caption{Annotated frames of the collected datasets containing a total of over $\numprint{70000}$ humans in different poses, in both optical and thermal imagery, and in a variety of different environments.}
 \label{pics:sample_own_data}
\end{figure}

To further increase the amount of training, validation, and test data we collected additional optical-infrared datasets.
The datasets are recorded with sensors mounted on the platforms listed in Fig.~\ref{fig:techpod} and \ref{fig:rega_drone}, at different locations and resembling realistic search-and-rescue missions.
A complete list of available datasets is shown in Tab.~\ref{tab:datasets_own}.
Along with all datasets we release the C++ based annotation tool that takes as input a dataset in the form of a rosbag \cite{Quigley2009} or video.
With the help of this tool, all humans appearing in the collected data have been annotated using upright rectangular bounding boxes.
During the labeling process, humans are assigned a unique ID. 
Every annotation contains an additional attribute for the human posture (\emph{upright}, \emph{sitting} or \emph{lying}) and one for occlusion (\emph{occluded}, \emph{not occluded}). 
Examples of the available annotated frames are illustrated in Fig.~\ref{pics:sample_own_data}.

%
%
%

%
\section{Experiments and Results}
\label{sec:Results}
%
\subsection{Experiment 1: Comparison of RetinaNet, YOLOv3 and Hand-crafted Detector}
\begin{figure}[htb]
\begin{minipage}[t]{0.45\textwidth}
\vspace{0pt}
\includegraphics[width=1\textwidth]{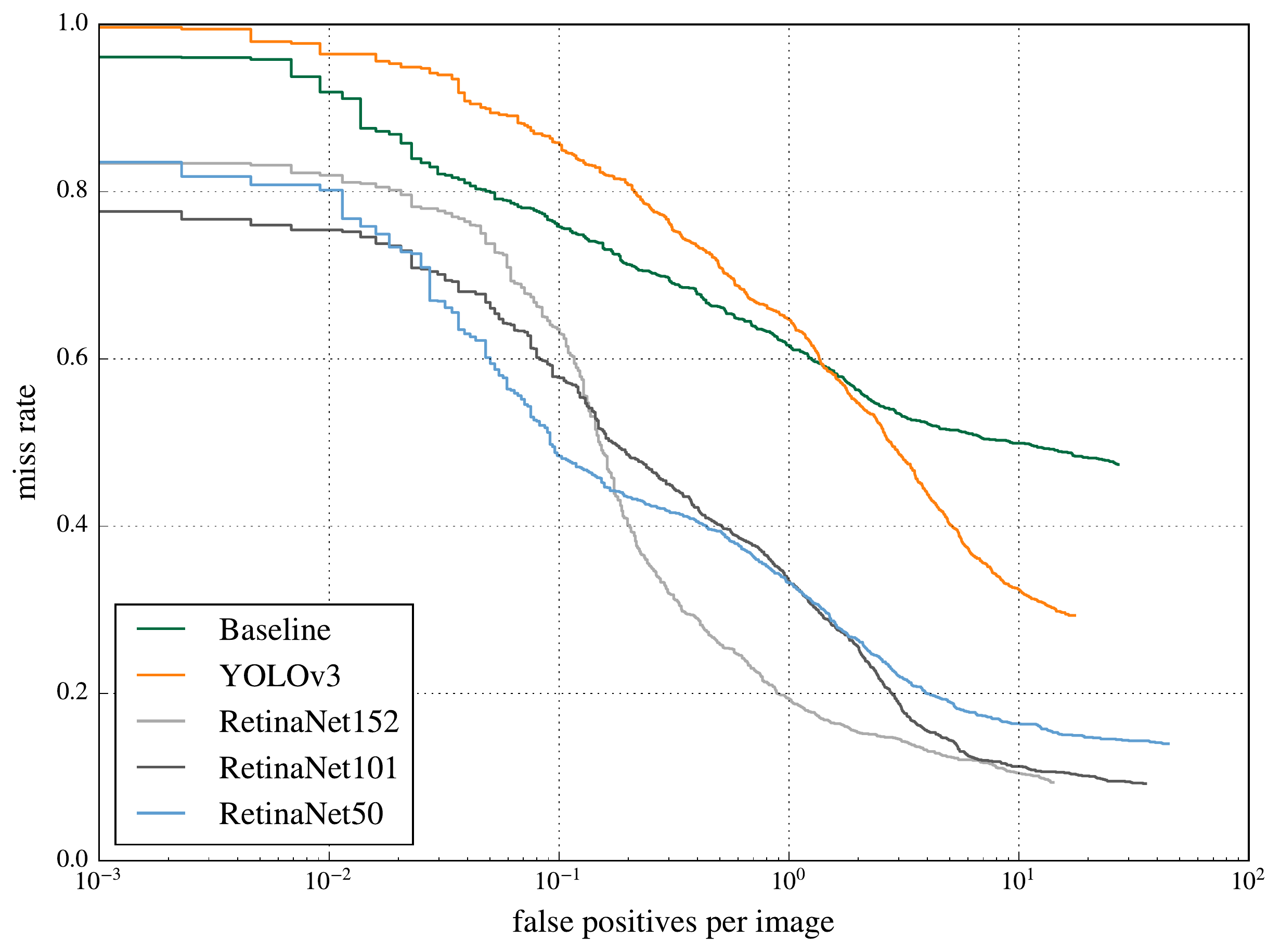}
\end{minipage}
\hfill
\begin{minipage}[t]{0.54\textwidth}
\vspace{0pt}
\resizebox{\textwidth}{!}{\begin{tabular}{cc}
\tiny RetinaNet152 & \tiny  \gls*{HOG} + \gls*{SVM} \cite{Kummerle2016}\\
\includegraphics[width=.49\linewidth]{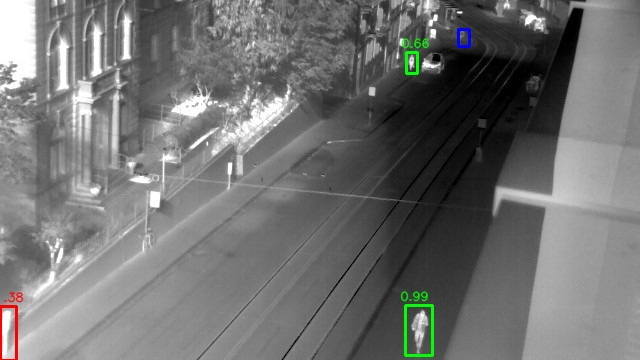} &
\includegraphics[width=.49\linewidth]{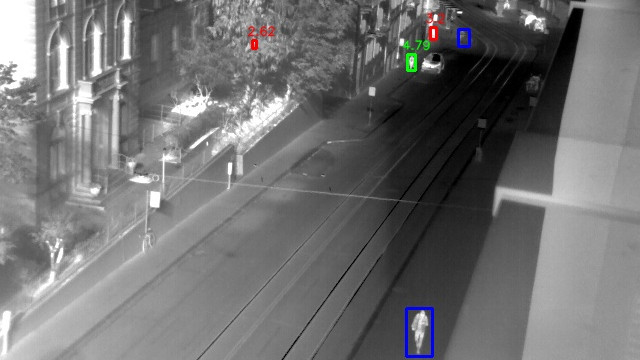}
\\
\includegraphics[width=.49\linewidth]{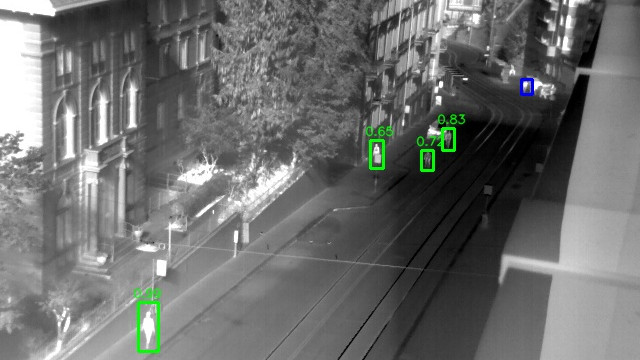} &
\includegraphics[width=.49\linewidth]{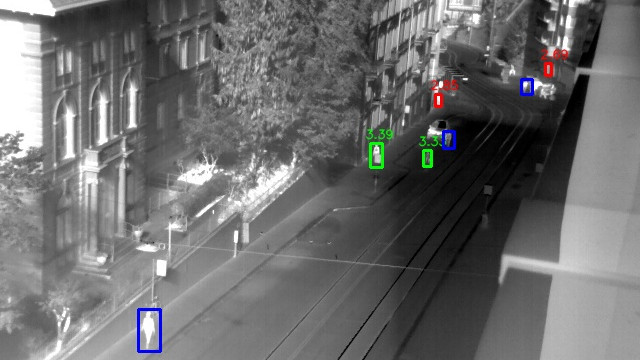}
\\
\end{tabular}}
\end{minipage}
\hspace*{\fill}
\caption{Thermal fppi/missrate curves on the \texttt{roof\_test} set illustrating the large performance improve of deep learning detectors.}
 \label{pics:dach_thermal}
\end{figure}

\begin{figure}[htb]
\begin{minipage}[t]{0.45\textwidth}
\vspace{0pt}
\includegraphics[width=1\textwidth]{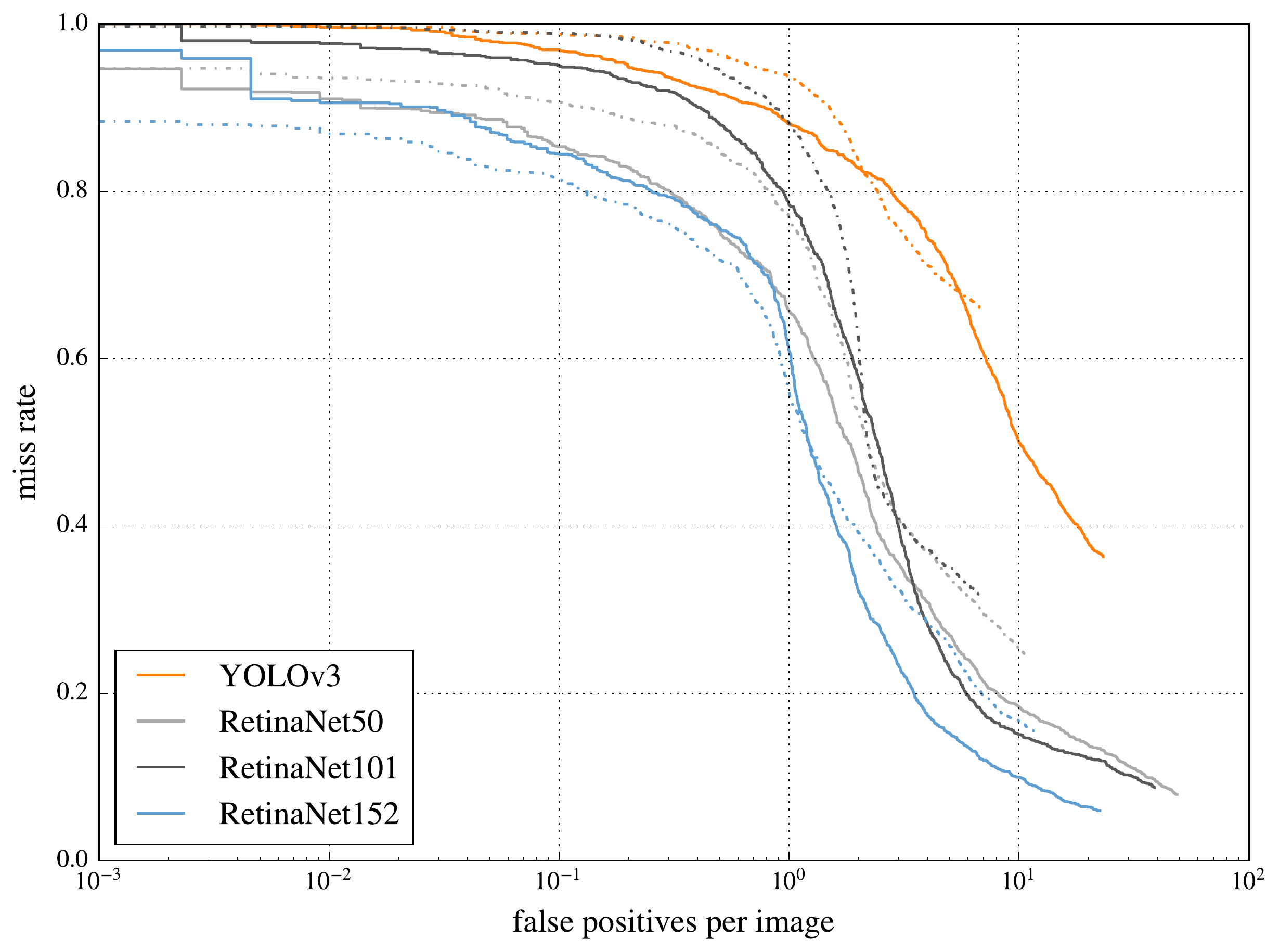}
\end{minipage}
\hfill
\begin{minipage}[t]{0.54\textwidth}
\vspace{0pt}
\resizebox{\textwidth}{!}{\begin{tabular}{cc}
\tiny RetinaNet152 logical AND & \tiny RetinaNet152 optical only\\
\includegraphics[width=.49\linewidth]{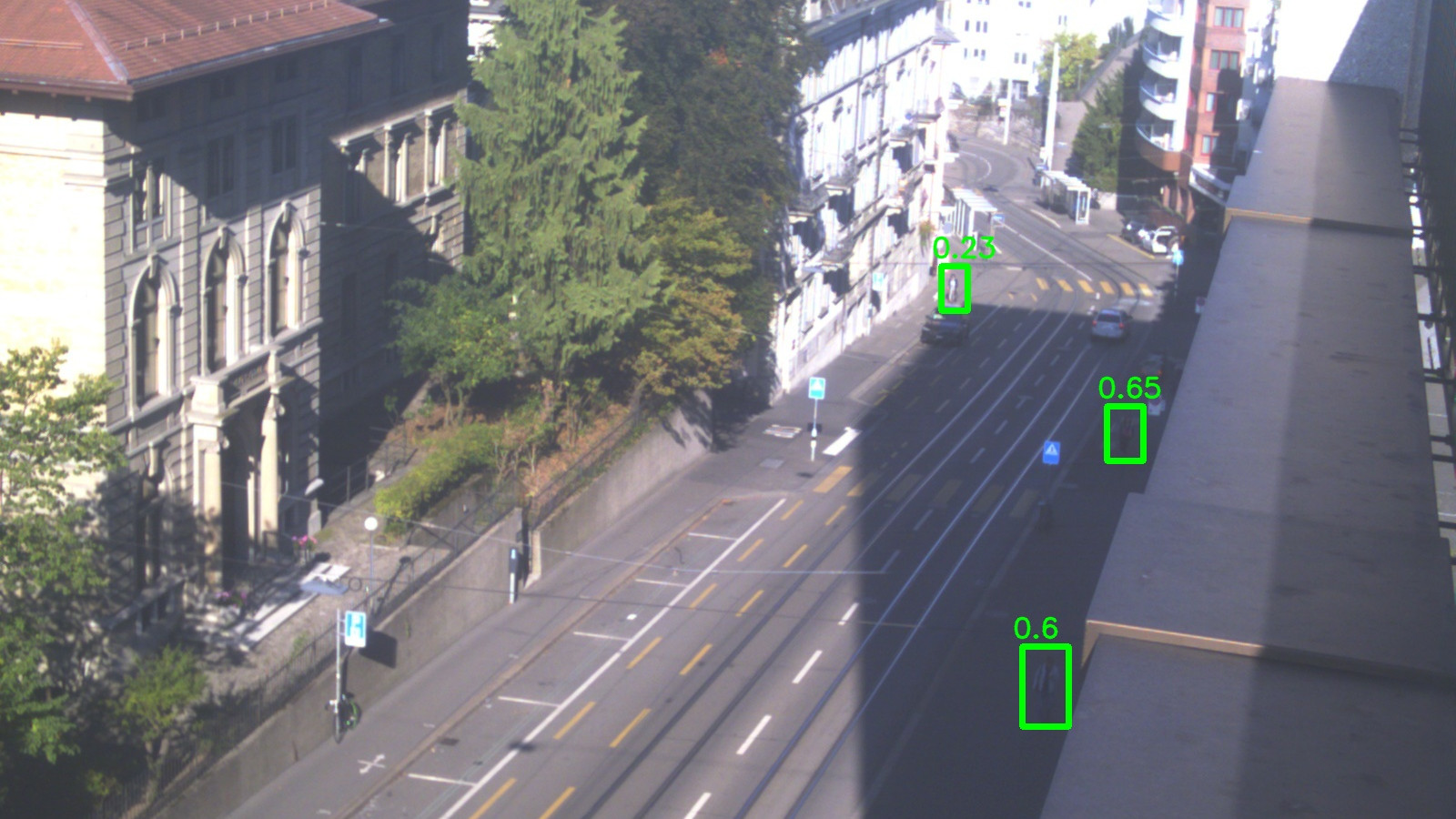} &
\includegraphics[width=.49\linewidth]{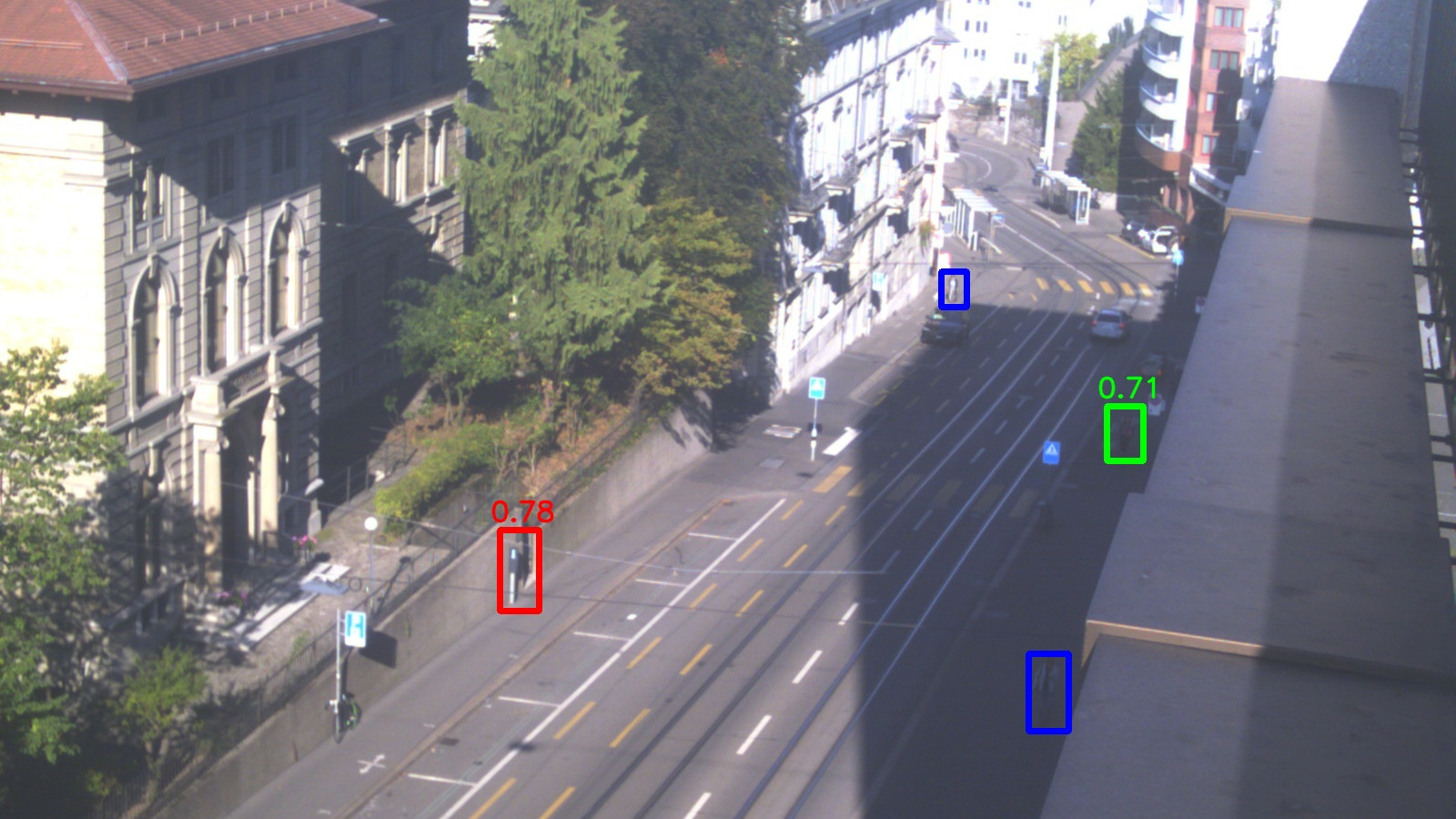}
\\
\includegraphics[width=.49\linewidth]{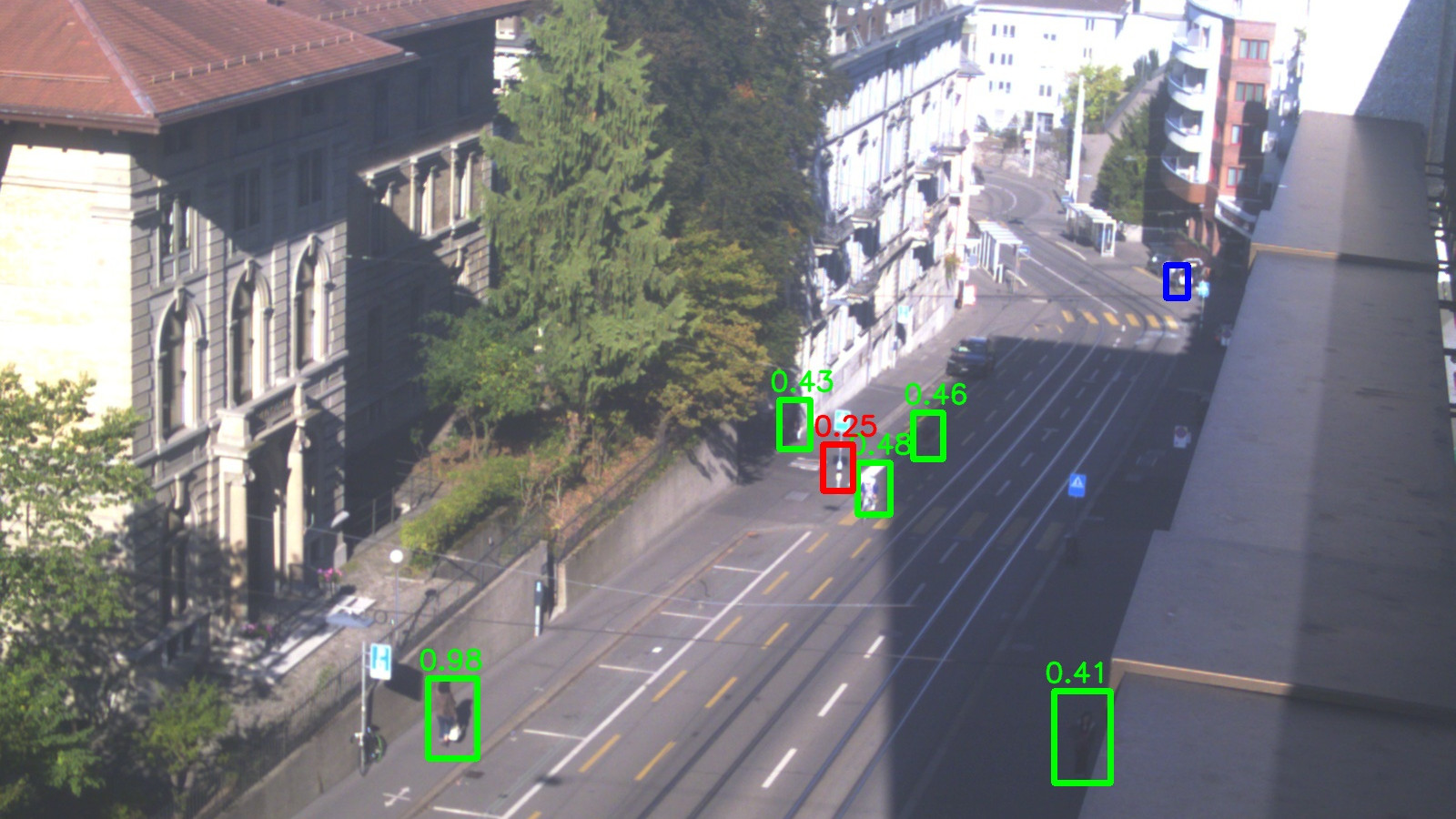} &
\includegraphics[width=.49\linewidth]{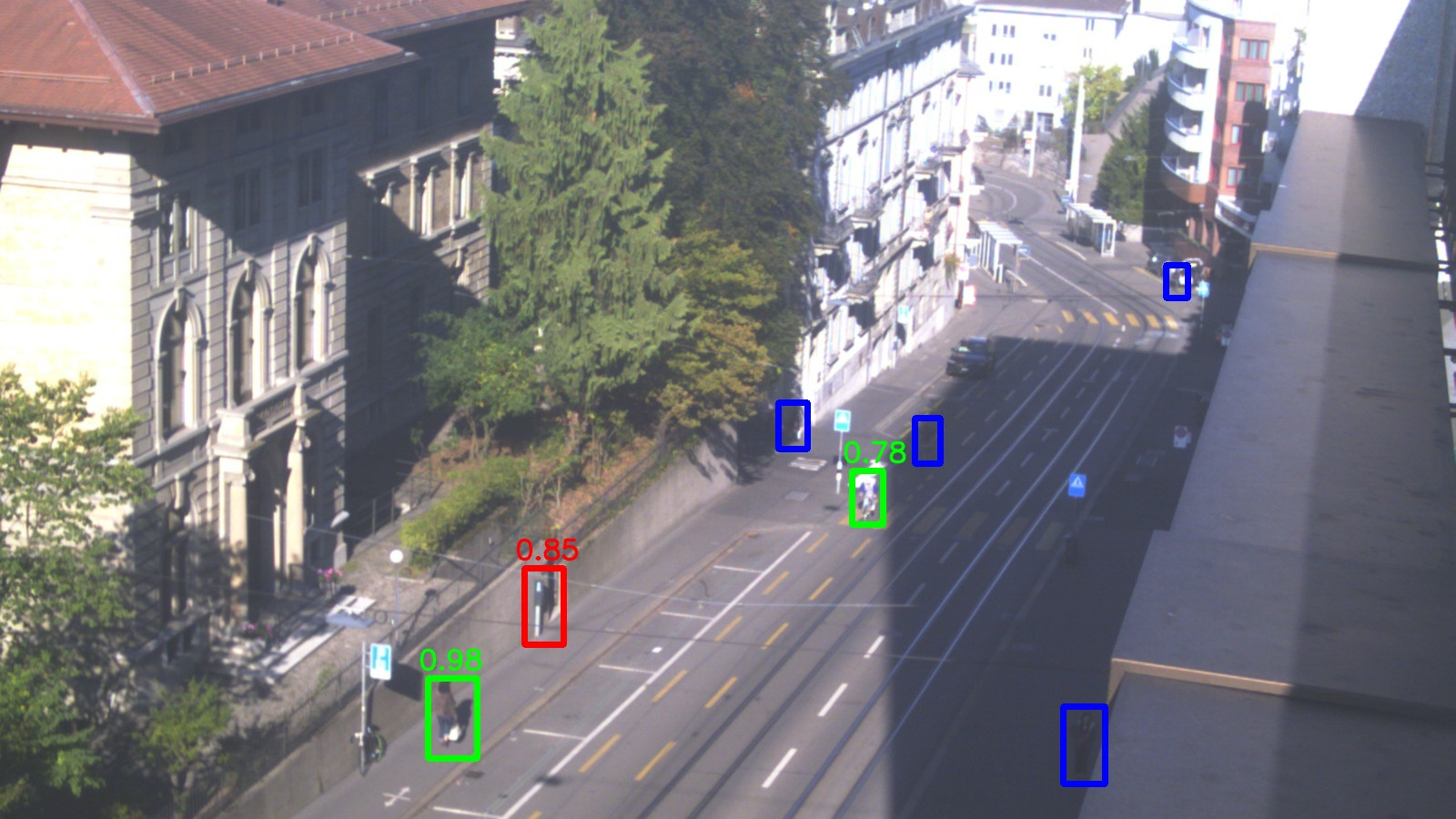}
\\
\end{tabular}}
\end{minipage}
\hspace*{\fill}
\caption{Optical fppi/missrate curves on the \texttt{roof\_test} set. Performance on pure optical information plotted as dashed lines and the improved performance using the logical OR merging scenario in bold. By combining optical and thermal information, both false positives and overall missrates are reduced.}
 \label{pics:dach_optical}
\end{figure}

In a first step, we compare the vanilla deep learning-based detectors to a hand-crafted human detection framework that uses \gls*{HOG} features together with a \gls*{SVM} classifier \cite{Kummerle2016} and that serves as a low baseline solution.
This reference pipeline, that was introduced in \cite{Kummerle2016}, detects humans in thermal imagery and then uses the corresponding optical image solely to reduce false positives.
%
%
The performance on the collected  \texttt{roof\_test} dataset, similar to the one by \cite{Kummerle2016}, is illustrated in Fig.~\ref{pics:dach_thermal} and \ref{pics:dach_optical}, and shows a significant improvement when using deep learning-based object detectors.
Furthermore, all of the RetinaNet variants vastly outperform the YOLOv3 framework.
%

\subsection{Experiment 2: RetinaNet evaluation in \gls*{SAR} scenario}
\begin{figure}[htb]
    \begin{minipage}[t]{0.45\linewidth}
   \includegraphics[width=\textwidth]{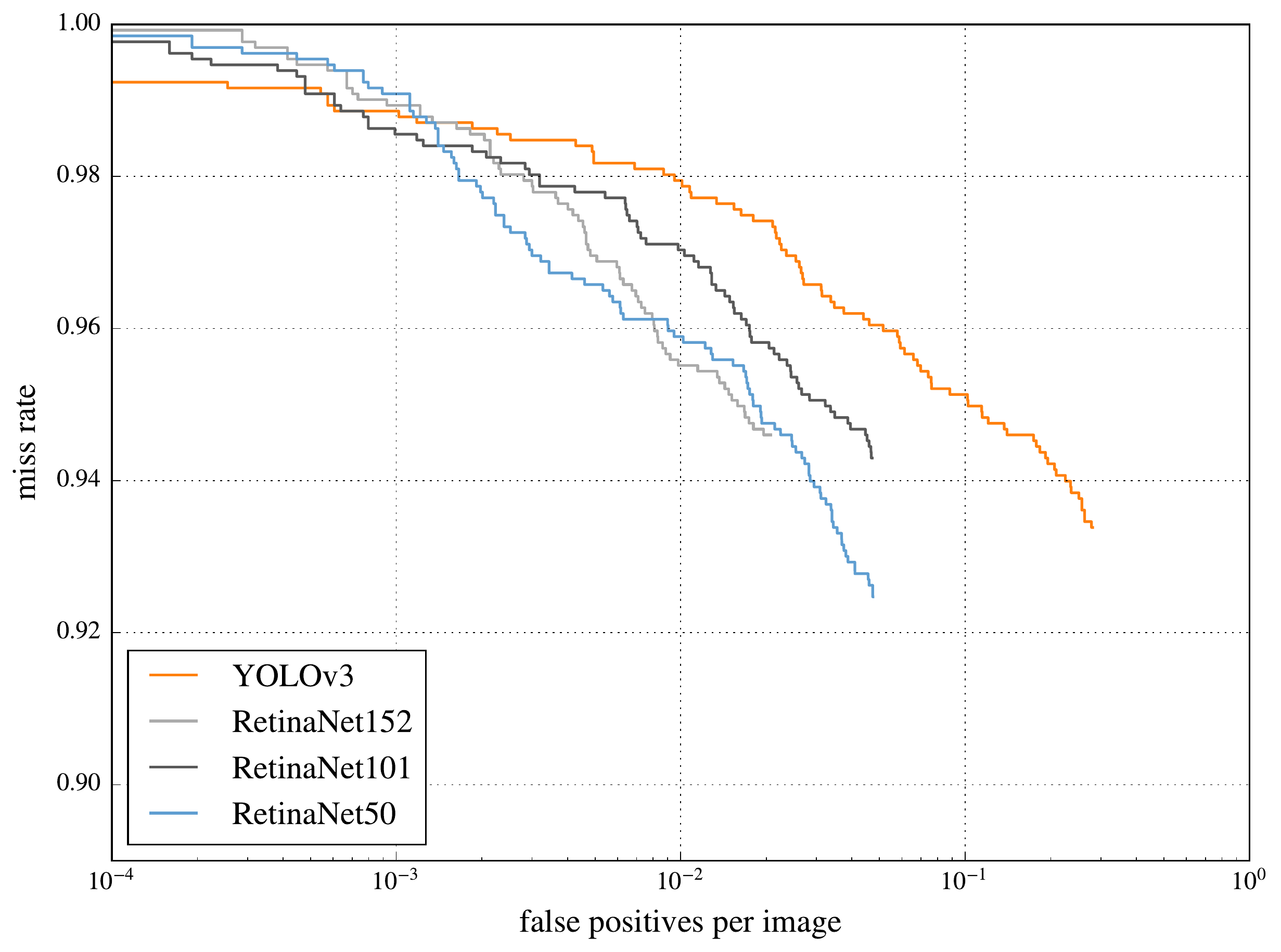}
   \caption{Performance of all networks on the total \texttt{field\_test} sets using standard networks without any changes on anchors, image resolution or any merging of IR-RGB information.}
   \label{pics:davos_fppi_missrate}
 \end{minipage}
 \hfill
 \begin{minipage}[t]{0.45\textwidth}
   \includegraphics[width=\textwidth]{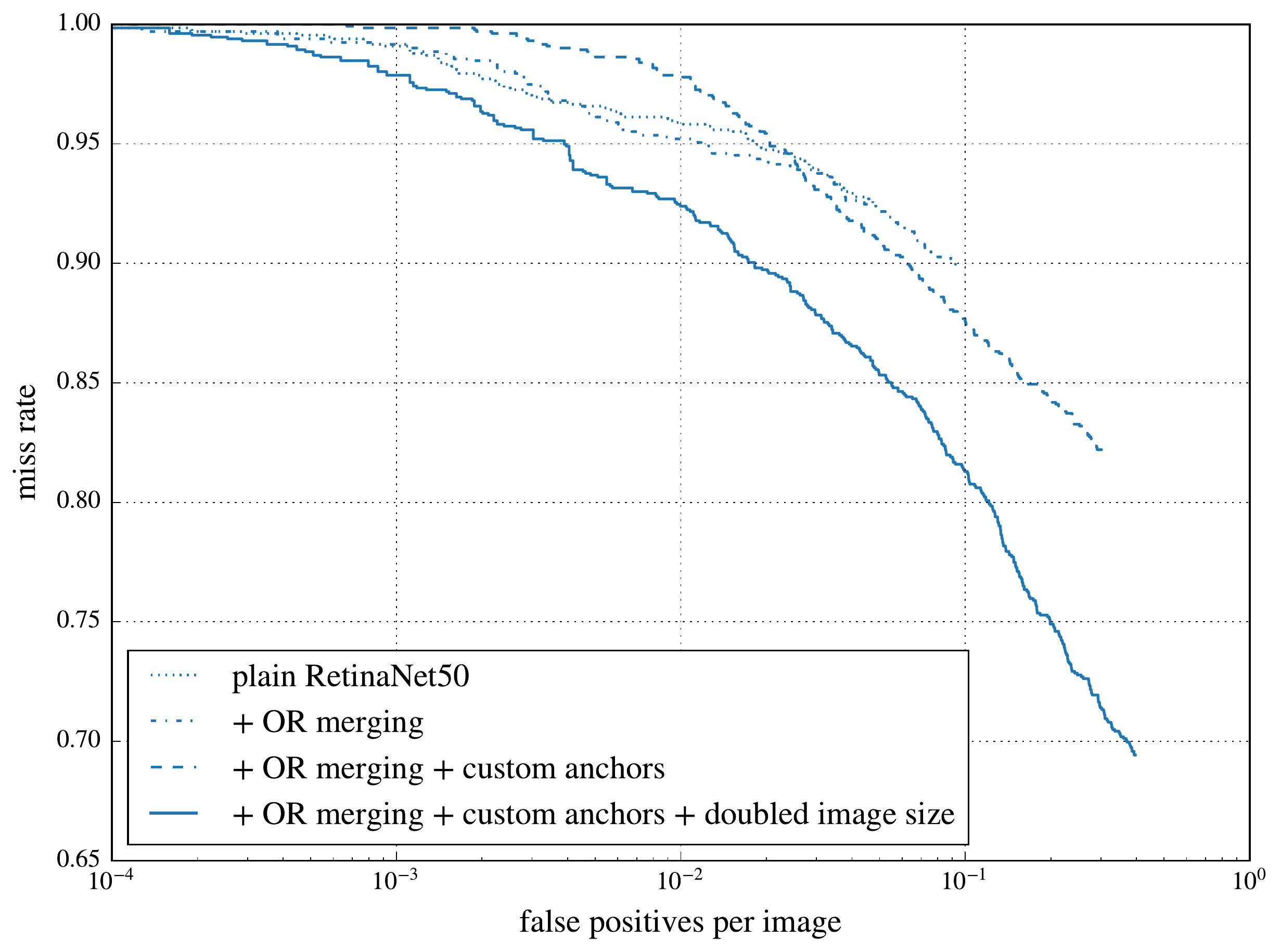}
   \caption{Performance on the total \texttt{field\_test} showing the gradual improvements when adding our modifications to the plain RetinaNet50 variants, as described in Sec.~\ref{subsec:network_customization}.}
   \label{pics:davos_2}
   \end{minipage}
    \end{figure}

%
%
%
In this section, the performance of the proposed human detection pipeline is thoroughly evaluated based on a dataset resembling a search-and-rescue scenario.
The \texttt{field\_test} datasets proved to be very challenging with a lot of very small human samples at different poses recorded from very high aerial views.
Occasional motion-blur due to the flight maneuvers and a lot of heated up rocks make the thermal imagery even more challenging.
The performance of our vanilla detectors on the total optical and thermal field evaluation datasets clearly emphasizes this:
Only a few percents of the total amount of human bounding boxes are detected.
Still, RetinaNet variants and especially the proposed RetinaNet50 network vastly outperforms YOLOv3 in the number of total detectable humans, as depicted in Fig.~\ref{pics:davos_fppi_missrate}.
%
 %
Finally, modifications to the best performing RetinaNet50 variant, as described in Sec.~\ref{subsec:network_customization}, show vast improvements for the \texttt{field\_test} dataset as illustrated in Fig.~\ref{pics:davos_2}.
While the logical OR merging scenario of optical and thermal detections already brings a performance improvement, the customized anchor bounding boxes vastly increase detection performance and boost up the total amount of detectable bounding boxes by over 7\%. Finally, doubling the image size during inference further increases this amount to a total of 70\% detected bounding boxes while making less than one false positive per image. This is an improvement of over 20\% when compared to the plain version of the original RetinaNet50 variant.
 
        \begin{figure}[htb]
    \begin{minipage}[t]{0.45\linewidth}
   \centering
   \includegraphics[width=\textwidth]{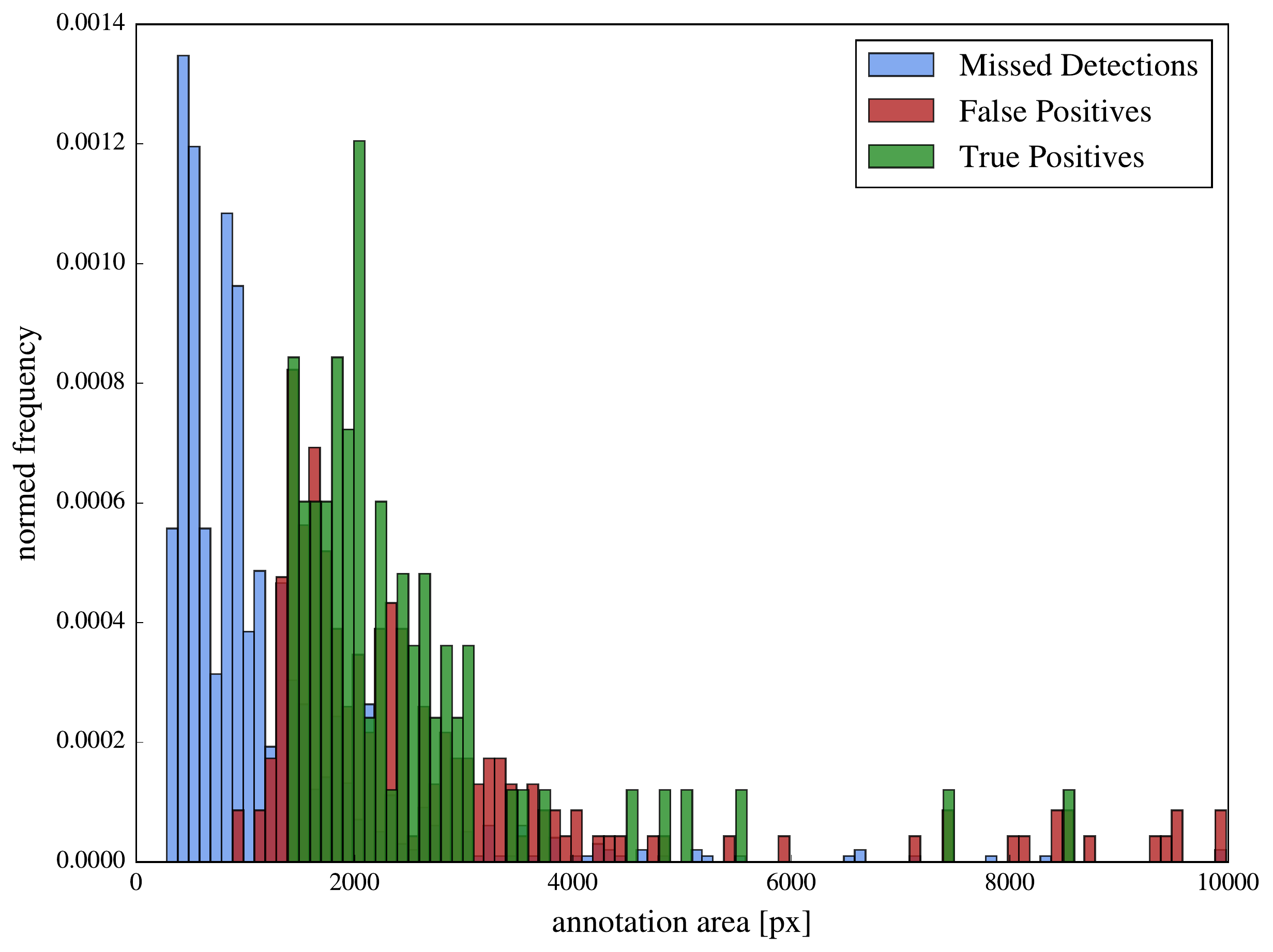}
   \caption{Bounding box size analysis of the best performing RetinaNet50 on the optical part of the \texttt{davos\_rega} test sets. RetinaNet is able to generate true positives at a bounding box size of around \unit[2000]{px} and fails to make any predictions with sizes below approximately \unit[1500]{px}.}
   \label{pics:size_analysis_retinanet_davos}
 \end{minipage}
 \hfill
 \begin{minipage}[t]{0.45\textwidth}
   \centering
   \includegraphics[width=\textwidth]{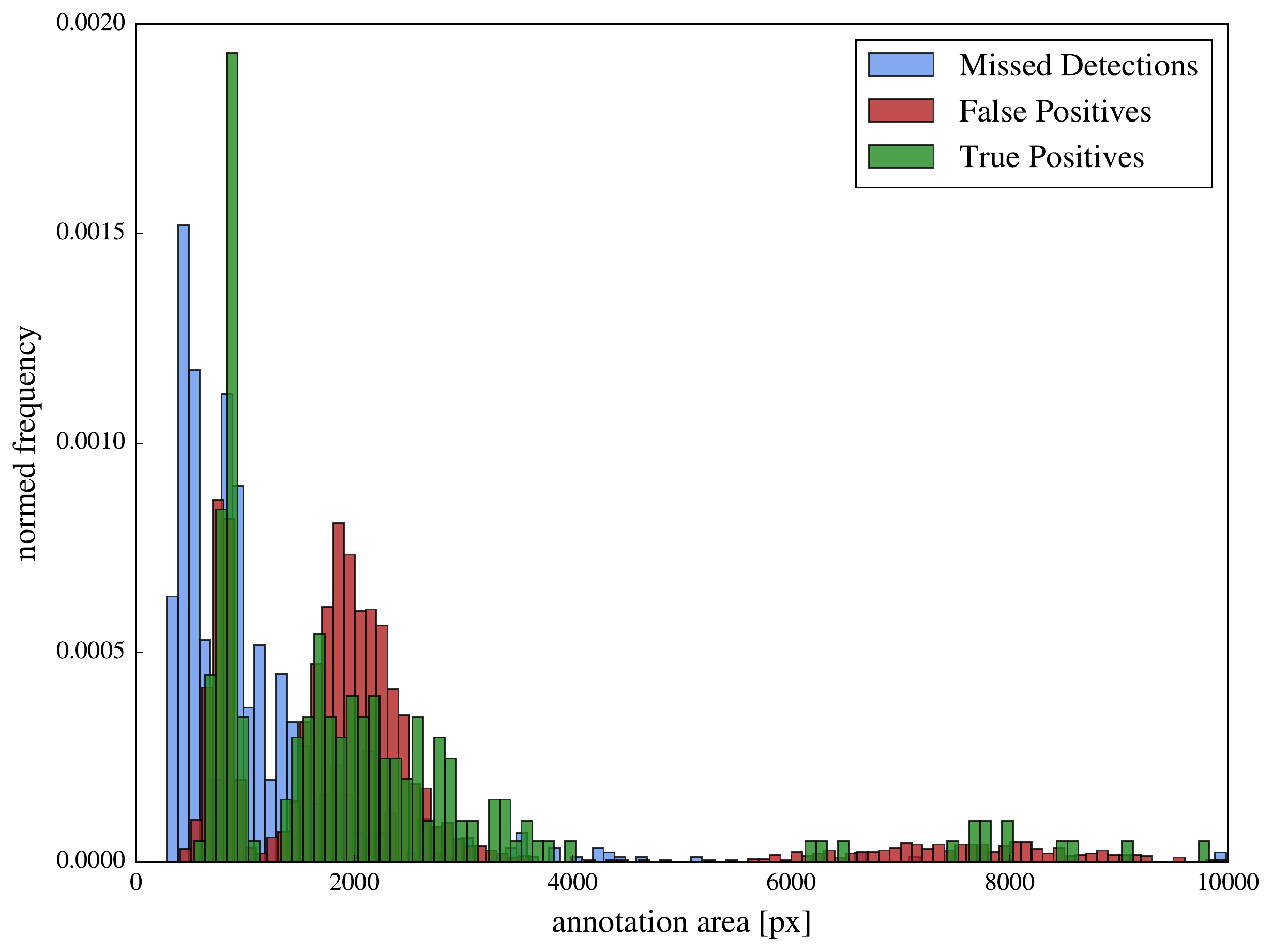}
   \caption{Bounding box size analysis of the best performing RetinaNet50 on the optical part of the \texttt{davos\_rega} test sets using custom anchors. RetinaNet now generates most of its true positive predictions at a bounding box size of around \unit[1000]{px} and is further able to make predictions across a wider range of sizes.}
   \label{pics:size_analysis_retinanet_davos_custom_anchors}
   \end{minipage}
 \end{figure}


%
\paragraph{Individual Human Detection}
A final evaluation investigates how well the pipeline detects human individuals:
Every human in our novel dataset is labeled with a unique ID.
Re-detections of the same individual can therefore be conveniently recognized.
For search and rescue scenarios, one person does not necessarily need to be detected in every single frame.
Instead, it is more important that an individual is detected at least once during the mission.
It is therefore informative to investigate how many of all the distinct individual human IDs are detected by the pipeline as illustrated in Fig.~\ref{pics:size_analysis_retinanet_davos} and \ref{pics:size_analysis_retinanet_davos_custom_anchors}.
By calculating the miss-rate of actual human IDs instead of single ground-truth bounding boxes, these final results show final human ID miss-rates of the best performing RetinaNet50 of less than 30\% while still making less than one false positive per image. 
This is a successful detection of over 70\% of all individual humans at least once.

\begin{figure}[htb]
   \centering
\resizebox{0.8\textwidth}{!}{\begin{tabular}{cc} 
 \includegraphics[width=0.4\linewidth]{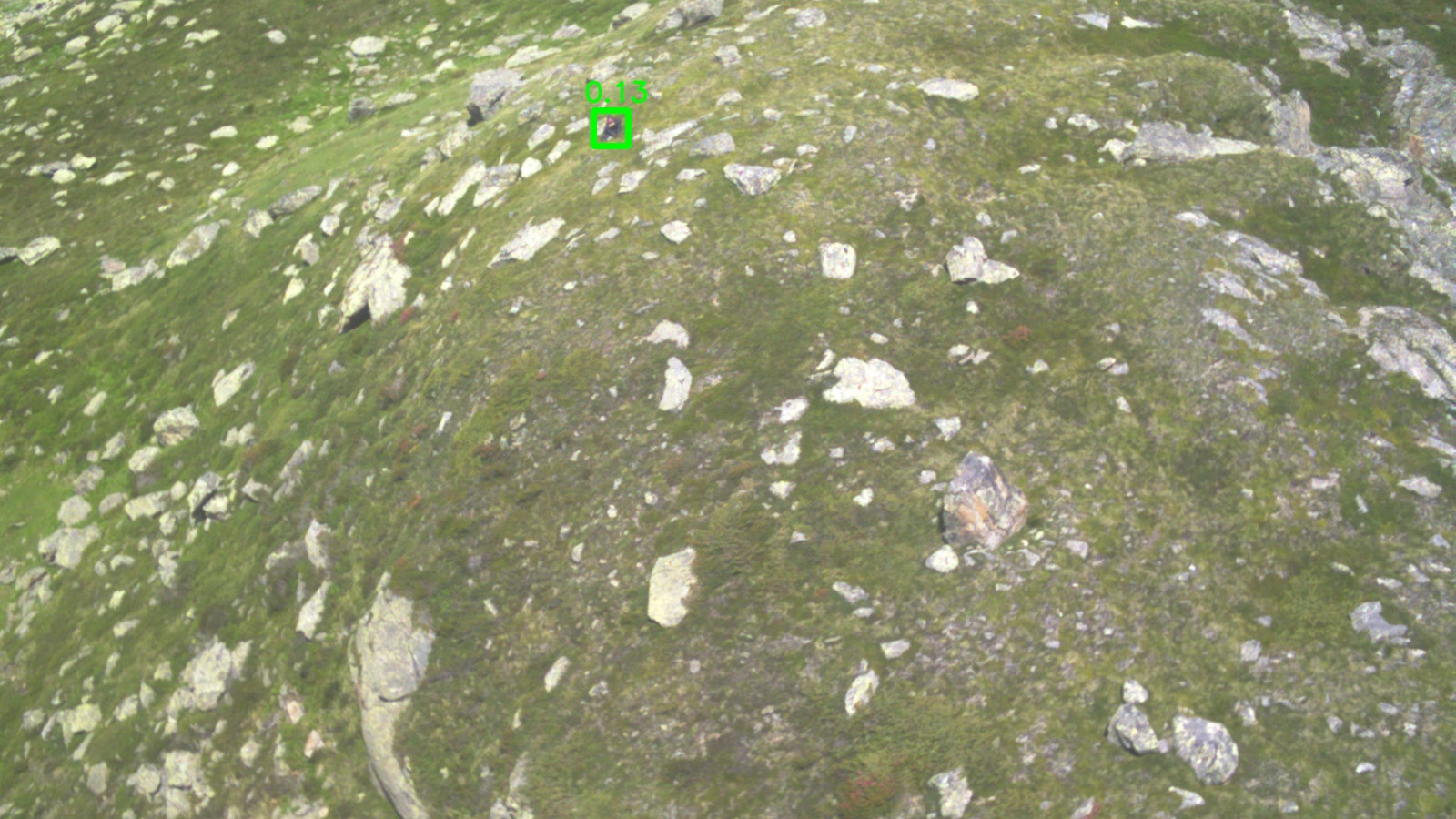} \hfill
 \includegraphics[width=0.4\linewidth]{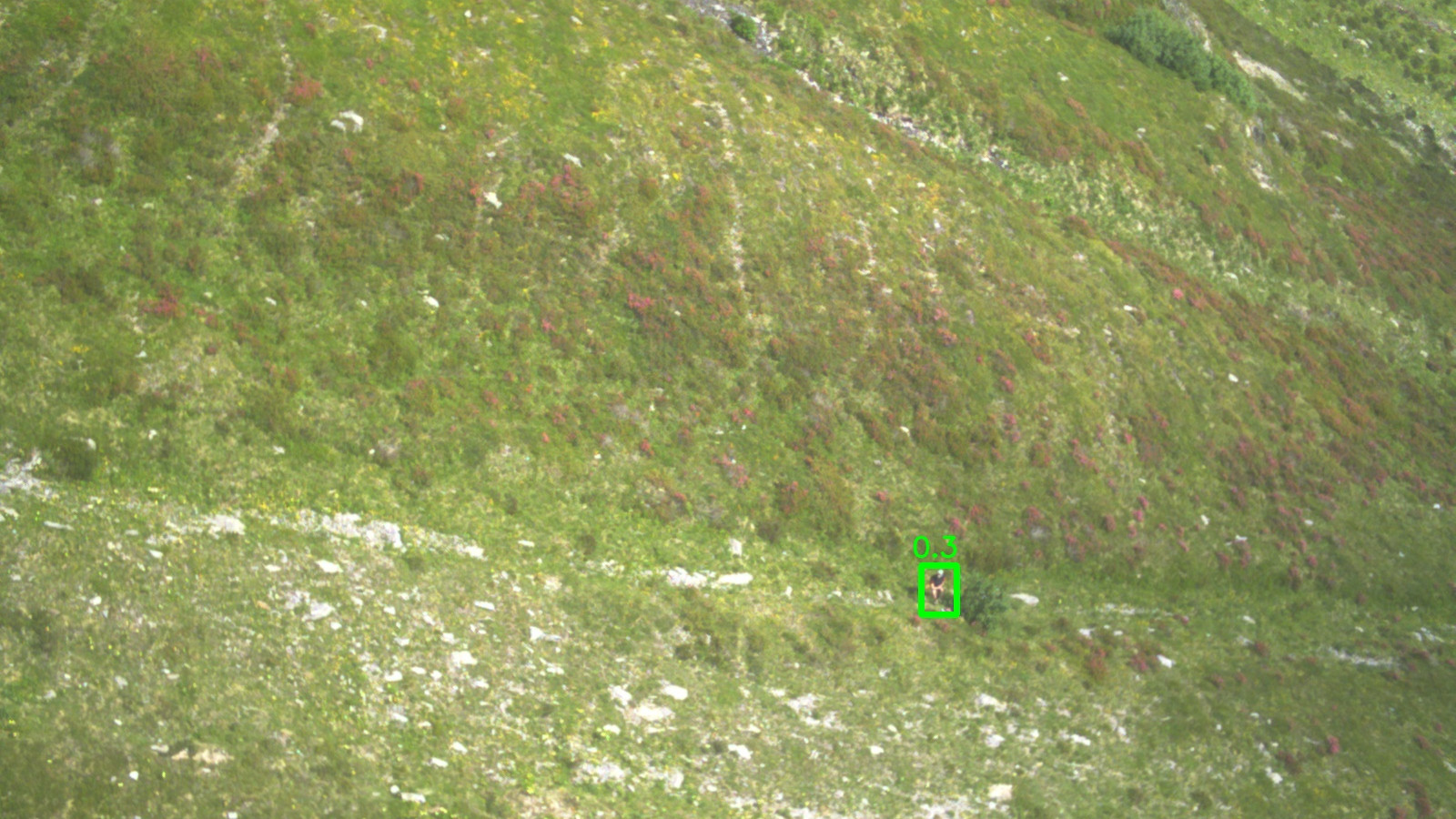} \\
 \includegraphics[width=0.4\linewidth]{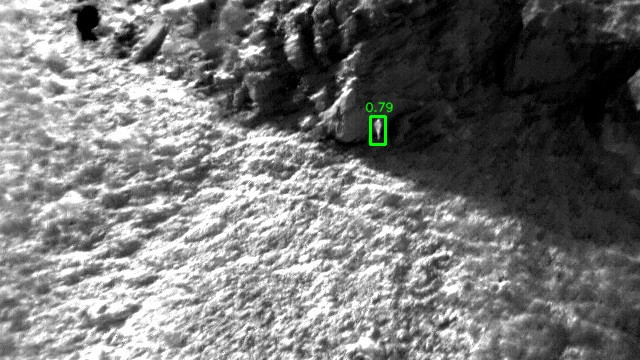} \hfill
 \includegraphics[width=0.4\linewidth]{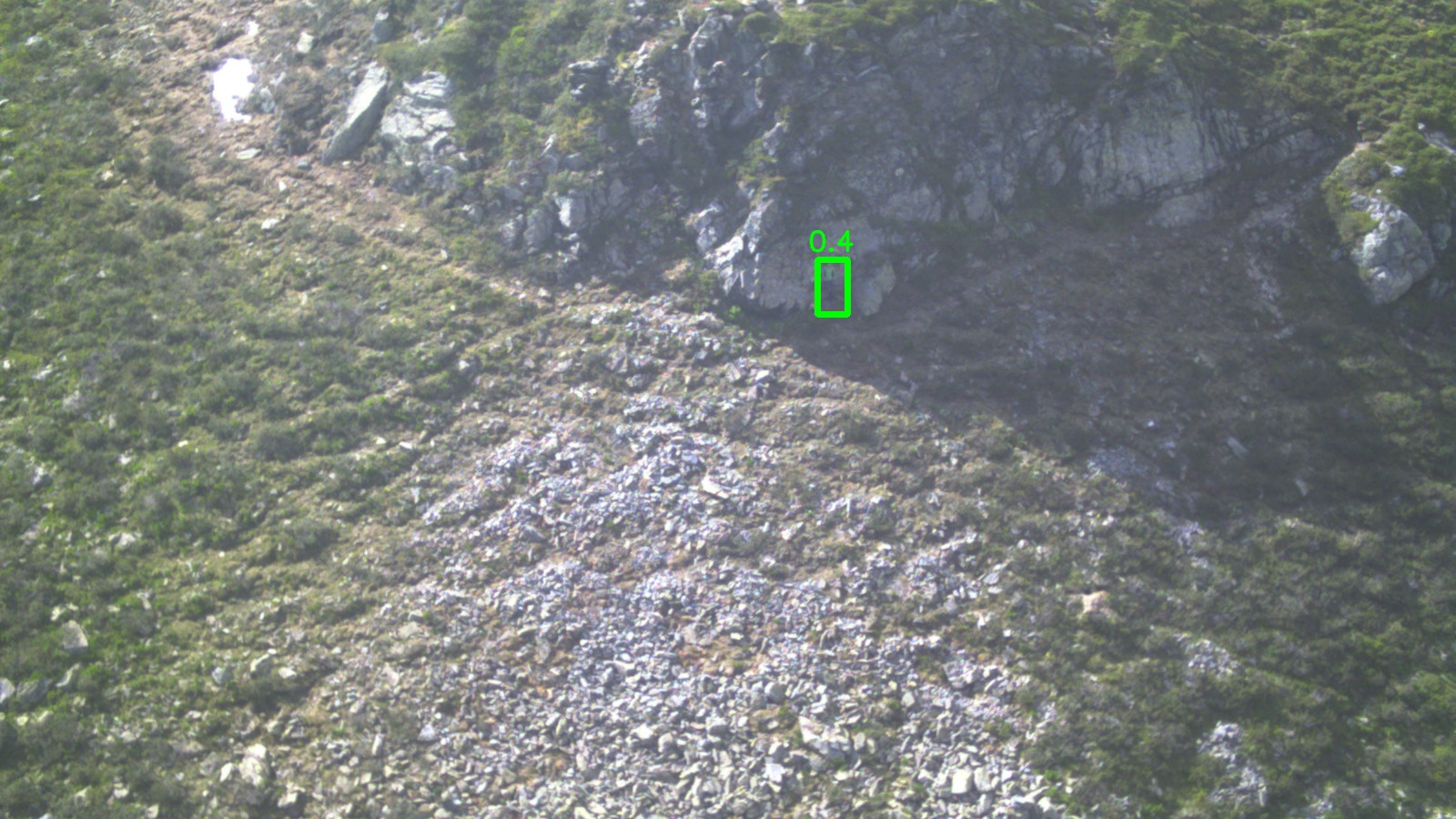} \\
  \includegraphics[width=0.4\linewidth]{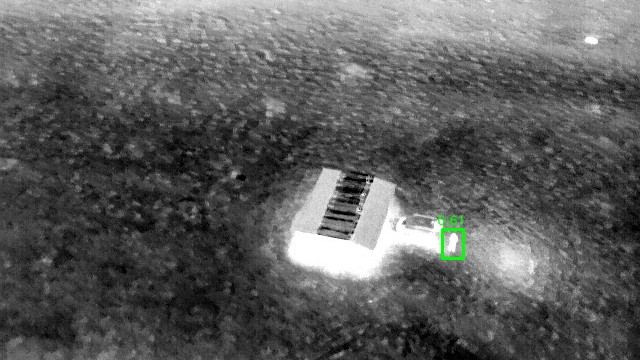} \hfill
  \includegraphics[width=0.4\linewidth]{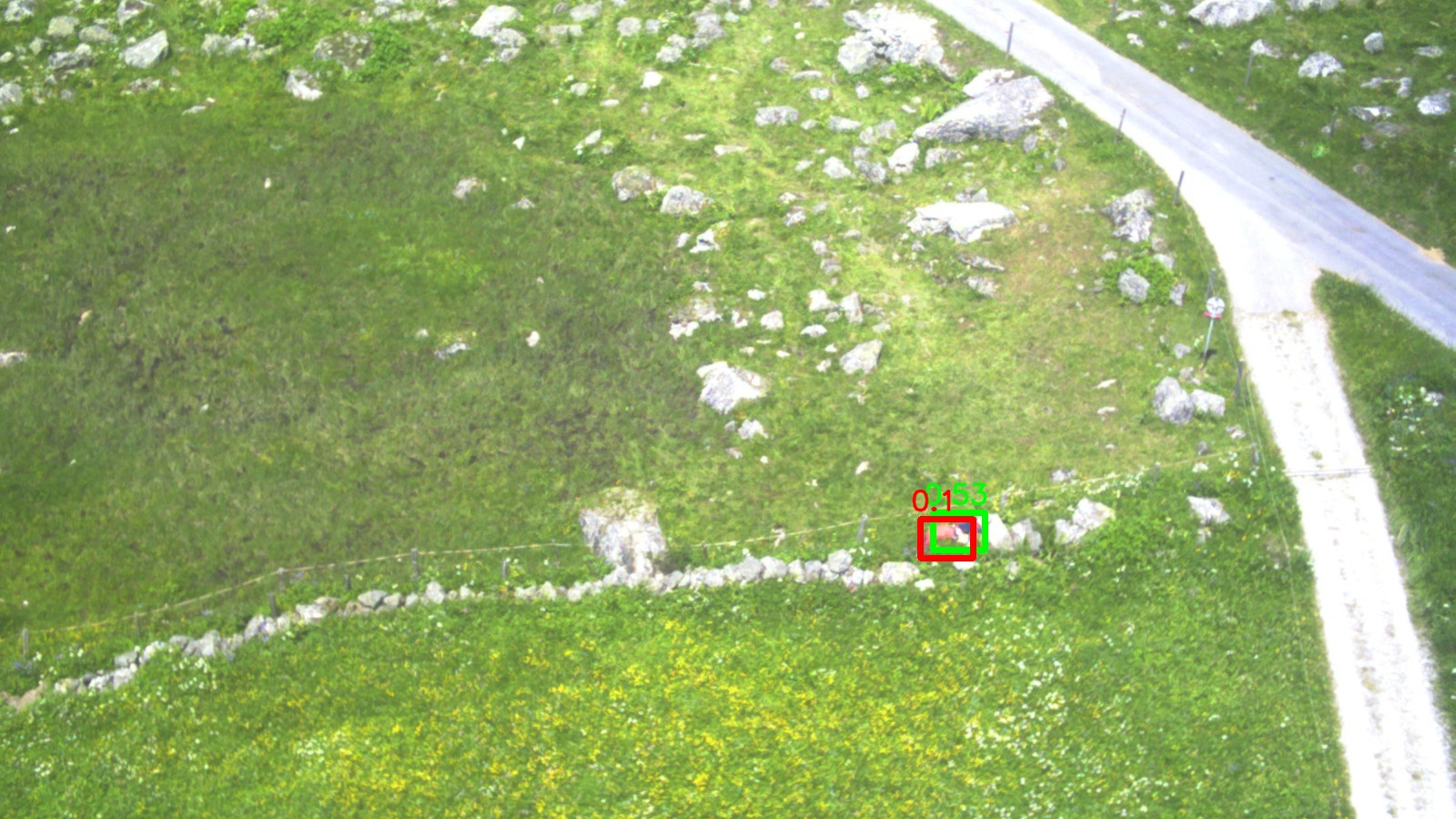} \\
  \includegraphics[width=0.4\linewidth]{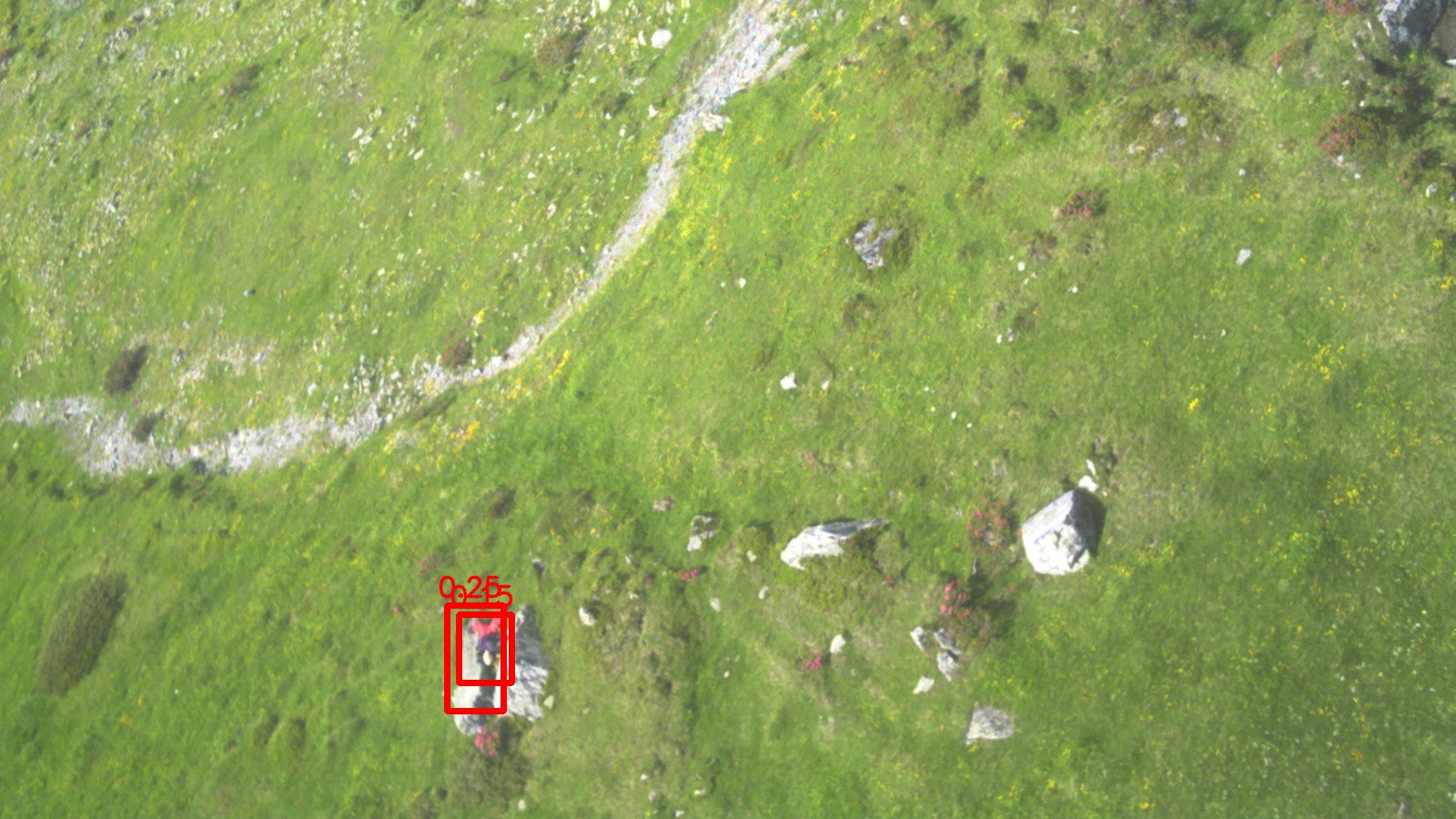} \hfill
 \includegraphics[width=0.4\linewidth]{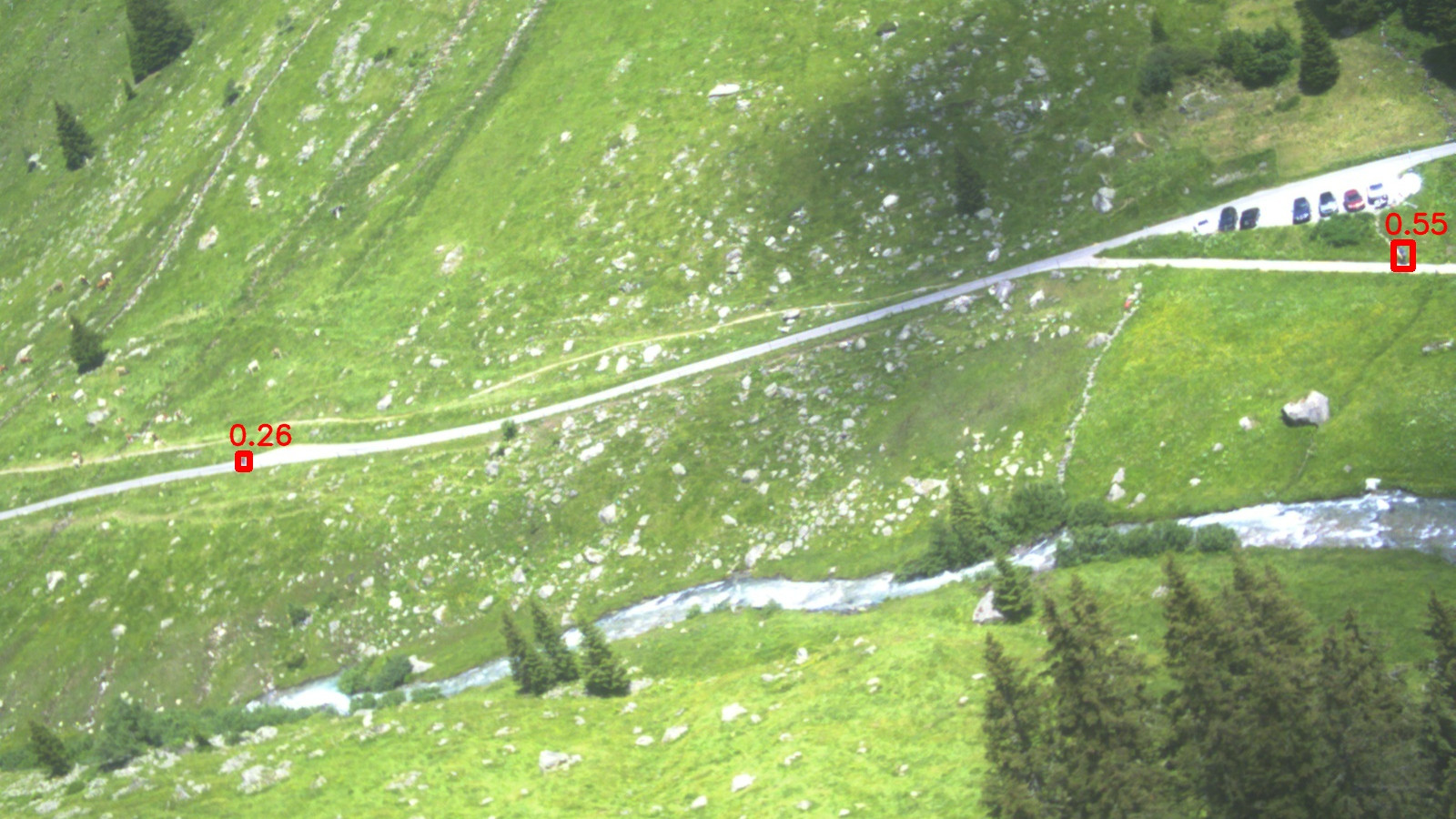}
 \end{tabular}}
 \caption{Qualitative samples. First row: Successfully detected humans, sitting and occluded. Second row: Working merging of optical and infrared image. Here, the human is easier to detect in the infrared spectrum. Third row: Night flight and detected mannequin that was placed before the flight. Fourth row: Correct detections by the network that have been missed during the manual labeling process.}
 \label{pics:qualitative_images}
\end{figure}

\paragraph{Qualitative samples}
Fig.~\ref{pics:qualitative_images} presents qualitative examples, including an example for the merging of the optical and infrared image and detected humans during a night flight.
 %

%
%
\section{Conclusion}
\label{sec:Conclusion}
This work presented our human detection framework that is able to detect, track, localize, and re-identify humans from \glspl*{UAV} with the help of an infrared and optical camera.
Based on a detailed evaluation, it can be concluded that the RetinaNet variants, in particular RetinaNet50, are superior to YOLOv3 and to a classical human detection pipeline that served as a lower baseline.
The major advantage of the RetinaNet architecture appears to be the focal loss that is capable of coping with drastic imbalances  between the number of foreground and background samples.
Moreover, customizing the anchors is crucial for the detection of humans seen from high altitudes.
Enlarged or higher resolution images further improve the detection performance.
Finally, the evaluation demonstrated that the logical OR merging scenario of both optical and thermal images helps to improve detection performance, especially for more challenging datasets like the introduced \texttt{field\_test} sets. 
Both the quantitative and the qualitative evaluation emphasize the conclusion that our novel pipeline indeed also works on a challenging real-world dataset successfully exploiting and combining information from both optical and thermal images. 
Given the performance of detecting individual human IDs and the final qualitative examples, it could even be questioned whether the human detection performance might already be surpassed by RetinaNet50 under certain circumstances. 
%

\section*{Acknowledgments}
The research leading to these results has received funding from the Swiss air rescue organization~\cite{rega}.
Furthermore, the authors wish to thank the following persons for their help: Thomas Mantel (Sensor triggering and calibration), Yves Allenspach (camera mounts), and Daniel Hentzen (circular calibration target for thermal camera and graphical user interface).
%

\FloatBarrier
\newpage
\section*{Appendix}
%
%
\begin{table}[htb]\footnotesize
  \caption{Recorded and labeled internal datasets.}
\begin{subtable}[t]{1\textwidth}
  \caption{Internal optical image sequences.}
\centering
\begin{tabular}{|l|l|l|l|l|l|l|l|}\hline
  \textbf{Sequence} 		& \multicolumn{2}{c|}{\textbf{Frames}} 	& \multicolumn{5}{c|}{\textbf{Annotations}} \\
  					& total 			& annotated 				& total 				& upright 			& sitting 		& lying 			& occluded 
  					 \\ \hline \hline
 roof\_new 			& \numprint{1900} 			& \numprint{1815} 					& \numprint{4748} 				& \numprint{4748} 				& \numprint{0} 				& \numprint{0} 				& \numprint{584}	\\
 roof\_old 			& \numprint{2898} 			& \numprint{2758} 					& \numprint{23165} 				& \numprint{23165} 				& \numprint{0}   			& \numprint{0}   			& \numprint{2396}			 \\
 roof\_test 			& \numprint{447} 			& \numprint{447} 					& \numprint{1462} 				& \numprint{1462} 				& \numprint{0}   			& \numprint{0}   			& \numprint{172}			 \\
 roof\_val 			& \numprint{231} 			& \numprint{231} 					& \numprint{991} 				& \numprint{991} 				& \numprint{0}   			& \numprint{0}   			& \numprint{258}			 \\
 davos\_rega\_01 	& \numprint{2553} 			& \numprint{86} 					& \numprint{159} 				& \numprint{116} 				& \numprint{12}  			& \numprint{31}  			& \numprint{21} 			\\
 davos\_rega\_02 	& \numprint{4806} 			& \numprint{360} 					& \numprint{543} 				& \numprint{374} 				& \numprint{0}   			& \numprint{169} 			& \numprint{37} 			 \\
 davos\_rega\_03 	& \numprint{5853} 			& \numprint{35} 					& \numprint{66} 				& \numprint{31}  				& \numprint{18}  			& \numprint{17}  			& \numprint{20} 			 \\
 davos\_rega\_04 	& \numprint{2962} 			& \numprint{203} 					& \numprint{322} 				& \numprint{166} 				& \numprint{3}   			& \numprint{153} 			& \numprint{29} 			 \\
 hinwil\_01 			& \numprint{2069} 			& \numprint{210} 					& \numprint{417} 				& \numprint{417} 				& \numprint{0}   			& \numprint{0}   			& \numprint{59} 			 \\
 hinwil\_02 			& \numprint{9028} 			& \numprint{1221} 					& \numprint{1994} 				& \numprint{1360}				& 530 			& \numprint{104} 			& \numprint{262}		\\
 solair 				& \numprint{7698} 			& \numprint{97} 					& \numprint{155} 				& \numprint{121} 				& \numprint{0}   			& \numprint{34}  			& \numprint{38} 			\\ \hline

\textbf{Total}		& \textbf{\numprint{40445}} & \textbf{\numprint{7463}}			& \textbf{\numprint{34022}} 	& \textbf{\numprint{32951}} 	& \textbf{\numprint{563}} 	& \textbf{\numprint{508}} 	& \textbf{\numprint{3876}} 	\\ \hline
 \end{tabular}
 \end{subtable}
 \\
 \vspace{12pt}

 \begin{subtable}[t]{1\textwidth} 
 \centering
  \caption{Internal infrared image sequences.}
\begin{tabular}{|l|l|l|l|l|l|l|l|}\hline
  \textbf{Sequence} 		& \multicolumn{2}{c|}{\textbf{Frames}} 	& \multicolumn{5}{c|}{\textbf{Annotations}} \\
  					& total 			& annotated 				& total 				& upright 			& sitting 		& lying 			& occluded 
  					 \\ \hline \hline
 roof\_new 			& \numprint{1834} 		& \numprint{1701} 					& \numprint{5752} 				& \numprint{5752} 				& \numprint{0} 				& \numprint{0} 				& \numprint{1825}			\\
 roof\_old 			& \numprint{1480} 			& \numprint{1316} 					& \numprint{6141}  				& \numprint{6141}  				& \numprint{0}   			& \numprint{0}   			& \numprint{477} 			\\
 roof\_test 			& \numprint{447} 			& \numprint{447} 					& \numprint{1421} 				& \numprint{1421} 				& \numprint{0}   			& \numprint{0}   			& \numprint{205}			\\
 roof\_val 			& \numprint{231} 			& \numprint{231} 					& \numprint{627} 				& \numprint{627} 				& \numprint{0}   			& \numprint{0}   			& \numprint{138}			\\
 davos\_old 			& \numprint{2204} 			& \numprint{664}					& \numprint{1004}				& \numprint{1004}				& \numprint{0}   			& \numprint{0}   			& \numprint{53} 			\\
 davos\_rega\_01 	& \numprint{2585} 			& \numprint{11} 					& \numprint{14}  				& \numprint{7}   				& \numprint{0}   			& \numprint{7}   			& \numprint{2}  			\\
 davos\_rega\_02 	& \numprint{4842} 			& \numprint{66}  					& \numprint{71}  				& \numprint{71}  				& \numprint{0}   			& \numprint{0}   			& \numprint{4}  			\\
 davos\_rega\_03 	& \numprint{5920} 			& \numprint{16} 					& \numprint{16} 				& \numprint{0}   				& \numprint{16}  			& \numprint{0}   			& \numprint{0}  			\\
 davos\_rega\_04 	& \numprint{2996} 			& \numprint{115} 					& \numprint{144} 				& \numprint{52}  				& \numprint{0}   			& \numprint{92} 			& \numprint{21} 			\\
 hinwil\_01 			& \numprint{2012} 			& \numprint{114} 					& \numprint{194} 				& \numprint{194} 				& \numprint{0}   			& \numprint{0}   			& \numprint{17} 			\\
 hinwil\_02 			& \numprint{730}  			& \numprint{15}   					& \numprint{15}   				& \numprint{0}   				& \numprint{15}  			& \numprint{0}   			& \numprint{2}  			\\
 rothenturm			& \numprint{37413}			& \numprint{9852}					& \numprint{21785}				& \numprint{21785}				& \numprint{0}   			& \numprint{0}   			& \numprint{1120}			\\
 solair 				& \numprint{27324}			& \numprint{0}  					& \numprint{0}   				& \numprint{0}   				& \numprint{0}   			& \numprint{0}   			& \numprint{0}  			\\
 tessin 				& \numprint{1011} 			& \numprint{43} 					& \numprint{44} 				& \numprint{27} 				& \numprint{0}   			& \numprint{17}   			& \numprint{1}			\\ \hline

\textbf{Total}		& \textbf{\numprint{91029}} & \textbf{\numprint{14591}}			& \textbf{\numprint{37228}} 	& \textbf{\numprint{37081}} 	& \textbf{\numprint{31}} 	& \textbf{\numprint{116}} 	& \textbf{\numprint{3856}} 	\\  \hline
 \end{tabular}
 \end{subtable}
  \label{tab:datasets_own}
\end{table}
\begin{table}[htb]
\caption{All external datasets used for training}
\begin{subtable}[t]{0.45\textwidth} 
\caption{Utilized external optical image sequences.}
\footnotesize\centering
\begin{tabular}{|l|l|l|l|l|}\hline
	 \textbf{Sequence}		& \multicolumn{2}{c|}{\textbf{Frames}} 		& \multicolumn{2}{c|}{\textbf{Annotations}} 	\\
  			& total 				& annotated 				& total 			& occluded 				\\ \hline  \hline
  			\multicolumn{5}{|c|}{Mini-drone \cite{minidrone} }\\ \hline
  			set10 \footnote{Original name: \newline Normal\_Static\_Night\_Empty\_1\_3\_1 (test)}	& \numprint{543}	& \numprint{542}	& \numprint{1207}	& \numprint{0} \\
		    set13 \footnote{Original name: \\ Normal\_Static\_Day\_Half\_0\_1\_1 (training)}	& \numprint{570}	& \numprint{569}	& \numprint{569}	& \numprint{0} \\ \hline \hline
  			\multicolumn{5}{|c|}{Stanford drone \cite{stanford} }\\ \hline
           	bookstore00	& \numprint{13335}	& \numprint{13335}	& \numprint{246158}	& \numprint{1494} 	\\
         	bookstore06	& \numprint{14558}	& \numprint{14305}	& \numprint{64944}	& \numprint{7090} 	\\
			coupa01		& \numprint{11966}	& \numprint{11966}	& \numprint{71136}	& \numprint{3726} 	\\
 			coupa02		& \numprint{11966}	& \numprint{11474}	& \numprint{66867}	& \numprint{3082} 	\\
 			gates07		& \numprint{2202}	& \numprint{2202}	& \numprint{14982}	& \numprint{365} 	\\
 			hyang02		& \numprint{12272}	& \numprint{12272}	& \numprint{172880}	& \numprint{4855} 	\\
 			hyang05		& \numprint{10648}	& \numprint{10648}	& \numprint{123521}	& \numprint{123} 	\\
 			hyang07		& \numprint{574}	& \numprint{574}	& \numprint{14637}	& \numprint{221} 	\\
 			hyang09		& \numprint{574}	& \numprint{574}	& \numprint{1930}	& \numprint{592} 	\\
 			hyang10		& \numprint{9928}	& \numprint{9928}	& \numprint{64228}	& \numprint{7845} 	\\
 			hyang12		& \numprint{9928}	& \numprint{9619}	& \numprint{39030}	& \numprint{2810} 	\\
			little00		& \numprint{1518}	& \numprint{1518}	& \numprint{24517}	& \numprint{508} 	\\
 			little01		& \numprint{14070}	& \numprint{13828} 	& \numprint{52399}	& \numprint{477} 	\\
 			quad03		& \numprint{509}	& \numprint{509}	& 2448	& \numprint{0} 		\\ \hline \hline
 			\multicolumn{5}{|c|}{UAV123 \cite{uav123}}\\ \hline
            bike01			& \numprint{553}	& \numprint{553}	& \numprint{553}	& \numprint{0} 	\\
 			person01		& \numprint{799}	& \numprint{799}	& \numprint{799}	& \numprint{0} 		\\
 			person02		& \numprint{2514}	& \numprint{2514}	& \numprint{2514}	& \numprint{0} 		\\
 			person03		& \numprint{643}	& \numprint{643}	& \numprint{643}	& \numprint{0} 		\\
 			person04		& \numprint{254}	& \numprint{254}	& \numprint{254}	& \numprint{0}		\\
 			person05		& \numprint{2101}	& \numprint{2101}	& \numprint{2101}	& \numprint{0} 		\\
 			person06		& \numprint{658} 	& \numprint{658}	& \numprint{658}	& \numprint{0} 		\\
 			person07		& \numprint{1943}	& \numprint{1873}	& \numprint{1873}	& \numprint{0} 		\\
 			person08		& \numprint{126}	& \numprint{126}	& \numprint{126}	& \numprint{0} 		\\
 			person10		& \numprint{582}	& \numprint{514}	& \numprint{514}	& \numprint{0} 		\\
 			person12		& \numprint{1621}	& \numprint{1548}	& \numprint{1548}	& \numprint{0}		\\
 			person13		& \numprint{155}	& \numprint{155}	& \numprint{155}	& \numprint{0} 		\\
 			person14		& \numprint{2034}	& \numprint{2034}	& \numprint{2034}	& \numprint{0} 		\\
 			person15		& \numprint{712}	& \numprint{712}	& \numprint{712}	& \numprint{0} 		\\
 			person16		& \numprint{1147}	& \numprint{1038}	& \numprint{1038}	& \numprint{0} 		\\
 			person17		& \numprint{1852}	& \numprint{1820}	& \numprint{1820}	& \numprint{0} 		\\
 			person22		& \numprint{24}		& \numprint{24}		& \numprint{24}		& \numprint{0} 		\\
 			person23		& \numprint{153}	& \numprint{153}	& \numprint{153}	& \numprint{0} 		\\
 			wakeboard02	& \numprint{733} 	& \numprint{733}	& \numprint{733}	& \numprint{0} 		\\
 			wakeboard03	& \numprint{748} 	& \numprint{748}	& \numprint{748}	& \numprint{0} 		\\
 			wakeboard04	& \numprint{586} 	& \numprint{586}	& \numprint{586}	& \numprint{0} 		\\
 			wakeboard05	& \numprint{758}	& \numprint{758}	& \numprint{758}	& \numprint{0} 		\\
 			wakeboard06	& \numprint{401} 	& \numprint{401}	& \numprint{401}	& \numprint{0} 		\\
 			wakeboard07	& \numprint{59}		& \numprint{59}		& \numprint{59}		& \numprint{0} 		\\
 			wakeboard08	& \numprint{321} 	& \numprint{321}	& \numprint{321}	& \numprint{0} 		\\ \hline \hline    
 			\multicolumn{1}{|r|}{\textbf{\numprint{41}}} & \textbf{\numprint{136638}} & \textbf{\numprint{134988}} & \textbf{\numprint{982578}} & \textbf{\numprint{33188}} \\ \hline       
\end{tabular}
 \label{tab:tabnefz}
\end{subtable}
\hfill
\begin{subtable}[t]{0.45\textwidth} 
\caption{Utilized external infrared image sequences.}      
\footnotesize\centering   
\begin{tabular}{|l|l|l|l|l|}\hline
	 \textbf{Sequence}		& \multicolumn{2}{c|}{\textbf{Frames}} 		& \multicolumn{2}{c|}{\textbf{Annotations}} 	\\ 
  			& total 				& annotated 				& total 			& occluded 				\\ \hline \hline
  			\multicolumn{5}{|c|}{ETH TIR \cite{eth_tir}}\\ \hline
         asl		    & \numprint{659}	& \numprint{659} 	& \numprint{1021} 	& \numprint{191}	\\
		 sempach06		& \numprint{600} 	& \numprint{413} 	& \numprint{413} 	& \numprint{0}		\\
 		 sempach07 		& \numprint{370} 	& \numprint{359} 	& \numprint{1391}	& \numprint{140}	\\
 		 sempach08 		& \numprint{634} 	& \numprint{634} 	& \numprint{1359} 	& \numprint{233}	\\
 		 sempach09 		& \numprint{576} 	& \numprint{576} 	& \numprint{982} 	& \numprint{59}		\\
 		 sempach10 		& \numprint{261} 	& \numprint{215} 	& \numprint{321} 	& \numprint{19}		\\
 		 sempach11 		& \numprint{197} 	& \numprint{192} 	& \numprint{707} 	& \numprint{113}	\\
 		 sempach12 		& \numprint{775} 	& \numprint{724} 	& \numprint{724} 	& \numprint{50}		\\  \hline\hline
 		 \multicolumn{5}{|c|}{OTCVBS 1 \cite{otcvbs1}}\\ \hline
         set01 			& \numprint{31} 	& \numprint{31} 	& \numprint{91} 	& \numprint{0}		\\
         set02 			& \numprint{28} 	& \numprint{28} 	& \numprint{100} 	& \numprint{0}		\\
		 set03 			& \numprint{23} 	& \numprint{23} 	& \numprint{101} 	& \numprint{0}		\\
		 set04 			& \numprint{18} 	& \numprint{18} 	& \numprint{109} 	& \numprint{0}		\\
		 set05 			& \numprint{23} 	& \numprint{23} 	& \numprint{101} 	& \numprint{0}		\\
		 set06 			& \numprint{18} 	& \numprint{18} 	& \numprint{97} 	& \numprint{0}		\\
 		 set07 			& \numprint{22} 	& \numprint{22} 	& \numprint{94} 	& \numprint{0}		\\
 		 set08 			& \numprint{24} 	& \numprint{24} 	& \numprint{99} 	& \numprint{0}		\\
 		 set09 			& \numprint{73} 	& \numprint{73} 	& \numprint{95} 	& \numprint{0}		\\
 		 set10 			& \numprint{24} 	& \numprint{24} 	& \numprint{97} 	& \numprint{0}		\\  \hline\hline
 		 \multicolumn{5}{|c|}{OTCVBS 11 \cite{otcvbs11}}\\ \hline
         set02 \footnote{Original name: set2/seq3/nuc.}			& \numprint{1273} 	& \numprint{1273} 	& \numprint{69841} 	& \numprint{0}		\\
         set03 \footnote{Original name: set2/set4/nuc.}			& \numprint{1131} 	& \numprint{1131} 	& \numprint{72686} 	& \numprint{0}		\\   \hline
         \multicolumn{5}{|c|}{PTB TIR \cite{ptb_tir}}\\ \hline
         stranger01 	& \numprint{95} 	& \numprint{95} 	& \numprint{95} 	& \numprint{0} \\
  		 stranger02		& \numprint{280} 	& \numprint{280} 	& \numprint{280} 	& \numprint{0} \\
		 stranger03		& \numprint{100} 	& \numprint{100} 	& \numprint{100} 	& \numprint{0} \\
	 	 walking 		& \numprint{315} 	& \numprint{315} 	& \numprint{315}	& \numprint{0} \\  \hline \hline
	 	 \multicolumn{5}{|c|}{VOT TIR \cite{vot_tir}}\\ \hline
         jacket 		& \numprint{1451} 	& \numprint{1451} 	& \numprint{1451} &	\numprint{178}		\\ \hline \hline
         \multicolumn{1}{|r|}{\textbf{\numprint{25}}} & \textbf{\numprint{9001}} & \textbf{\numprint{8701}} & \textbf{\numprint{152670}} & \textbf{\numprint{983}} \\  \hline
         \end{tabular}
 \label{tab:tabnefz}
\end{subtable}
\end{table}

\FloatBarrier
\bibliography{lib_complete.bib}
\bibliographystyle{abbrvnat}

\end{document}